\documentclass{article}

\usepackage[preprint]{neurips_2025}

\usepackage[utf8]{inputenc} % allow utf-8 input
\usepackage[T1]{fontenc}    % use 8-bit T1 fonts
\usepackage[pagebackref=true,breaklinks=true,colorlinks,bookmarks=false,citecolor=blue,linkcolor=blue]{hyperref}
\usepackage{url}            % simple URL typesetting
\usepackage{booktabs}       % professional-quality tables
\usepackage{amsfonts}       % blackboard math symbols
\usepackage{nicefrac}       % compact symbols for 1/2, etc.
\usepackage{microtype}      % microtypography
\usepackage{xcolor}         % colors
\usepackage{multicol}
\usepackage{multirow}
\usepackage{booktabs}
\usepackage{caption}
\usepackage{subcaption}
\usepackage{tabularx}
\usepackage{newfloat}
\usepackage{listings}
\usepackage{tcolorbox}
\usepackage{algorithm}
\usepackage{algorithmic}
\usepackage{colortbl}
\usepackage{lscape}
\usepackage{graphicx}
\usepackage{pifont}
\usepackage{subcaption} 
\usepackage{wrapfig}
\usepackage{floatflt}
\usepackage{tcolorbox}
\usepackage{marvosym}
\usepackage{afterpage}

\newcommand{\ourbench}[0]{Video-Holmes}
\definecolor{color1}{HTML}{ECF4F9}
\definecolor{color2}{HTML}{FFF1E0}
\definecolor{color3}{HTML}{ECF4E9}

\definecolor{GenColor}{HTML}{6D339C}
\definecolor{RepColor}{HTML}{285D9B}

\definecolor{Gray}{gray}{0.95}
\definecolor{LightBlue}{RGB}{236,244,249}
\newcommand{\CC}[1]{\cellcolor{LightBlue}}
\newcommand{\RC}[1]{\rowcolor{LightBlue}}
\definecolor{mydeepgreen}{RGB}{0,100,0} % 深绿色
\definecolor{deepred}{RGB}{139,0,0} % 这是一个深红色的RGB值

\title{\ourbench: Can MLLM Think like Holmes for Complex Video Reasoning?}

\author{
\vspace{2mm}
Junhao Cheng\textsuperscript{1,2}, 
	  ~Yuying Ge\textsuperscript{1,\Letter},
        ~Teng Wang\textsuperscript{1,\Letter},
        ~Yixiao Ge\textsuperscript{1},
        ~Jing Liao \textsuperscript{2},
        ~Ying Shan\textsuperscript{1} \\ 
\vspace{2mm}
	$^1$ARC Lab, Tencent PCG\qquad $^2$City University of Hong Kong \\
    \url{https://video-holmes.github.io/Page.github.io/}
}

\begin{document}

\maketitle

\begin{figure*}[h!]
	\centering
	\includegraphics[width=\textwidth]{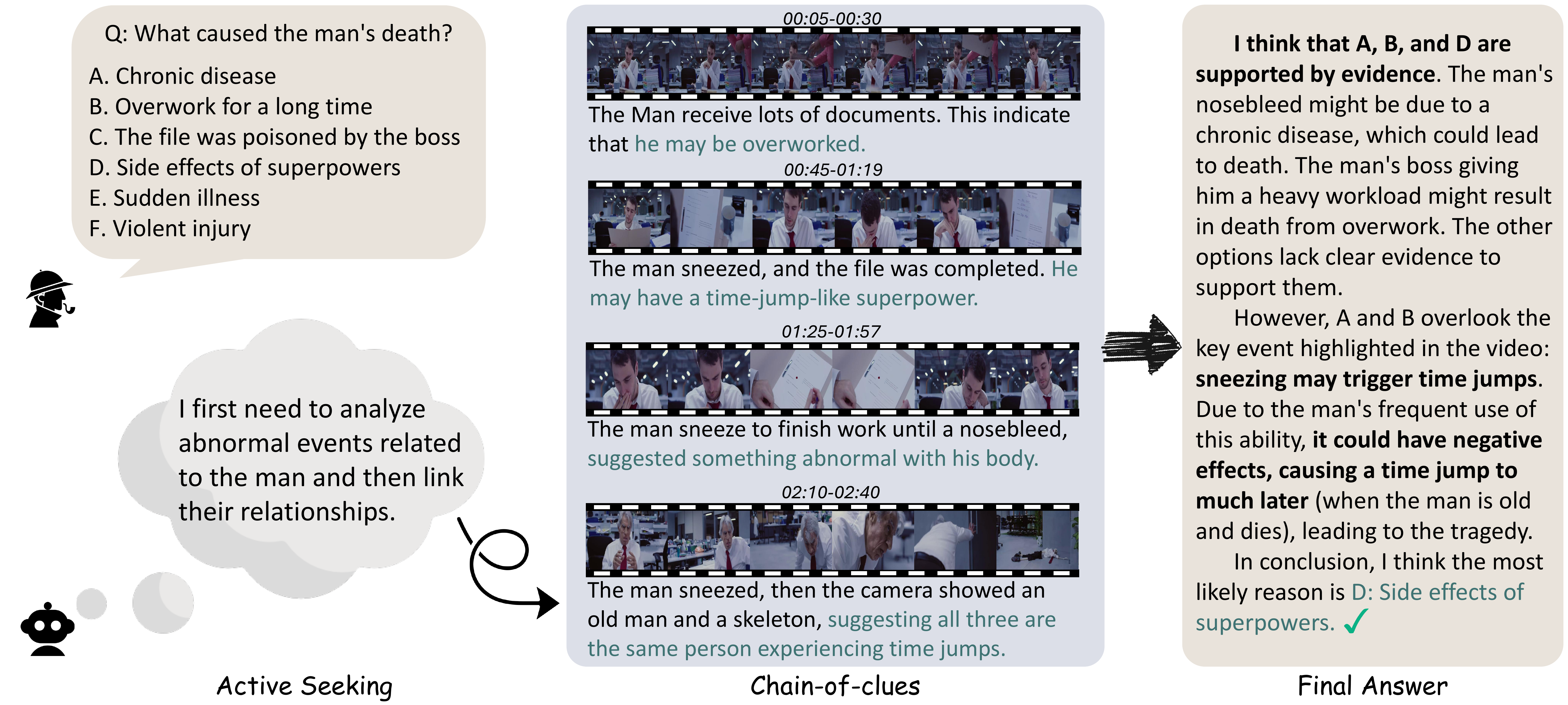}
	\caption{An example of Video-Holmes. Models are required to actively locate and connect multiple relevant visual clues scattered across different video segments to render the final answer.}
    %\vspace{-1em}
\label{fig: teasor}
\end{figure*}

\begin{abstract}
Recent advances in Chain-of-Thought (CoT) reasoning and reinforcement learning (RL) post-training have been reported to enhance video reasoning capabilities of multimodal large language models (MLLMs). This progress naturally raises a question: \textit{can these models perform complex video reasoning in a manner comparable to human experts?} However, existing video benchmarks primarily evaluate visual perception and grounding abilities, with questions that can be answered based on explicit prompts or isolated visual cues (e.g., ``What is the woman wearing?''). Such benchmarks do not fully capture the intricacies of real-world reasoning, where humans must actively search for, integrate, and analyze multiple clues before reaching a conclusion. Empirical results show that even models with advanced thinking abilities achieve only marginal gains (e.g., from 68.3\% to 69.4\%) on these benchmarks, which raises doubts about the extent to which these tasks require genuine reasoning.
To address this issue, we present \textbf{Video-Holmes}, a benchmark inspired by the reasoning process of Sherlock Holmes, designed to evaluate the complex video reasoning capabilities of MLLMs. Video-Holmes consists of 1,837 questions derived from 270 manually annotated suspense short films, which spans seven carefully designed tasks. Each task is constructed by first identifying key events and causal relationships within films, and then designing questions that require models to actively locate and connect multiple relevant visual clues scattered across different video segments. We conduct a detailed analysis of model reasoning processes, examining the factors that lead to both correct and incorrect answers. Our comprehensive evaluation of state-of-the-art MLLMs reveals that, while these models generally excel at visual perception, they encounter substantial difficulties with integrating information and often miss critical clues. For example, the best-performing model, Gemini-2.5-Pro, achieves an accuracy of only 45\%, with most models scoring below 40\%. We aim that Video-Holmes can serve as a \textit{``Holmes-test''} for multimodal reasoning, motivating models to reason more like humans and emphasizing the ongoing challenges in this field. The benchmark is released in \url{https://github.com/TencentARC/Video-Holmes}.

\end{abstract}

\section{Introduction}

The development of CoT reasoning~\cite{wei2022chain} and RL post-training strategies~\cite{shao2024deepseekmath} have contributed to significant improvements in the reasoning abilities of LLMs~\cite{guo2025deepseek,o1,o3}. By generating human-like reasoning steps, these models have shown strong performance in addressing complex reasoning tasks. Furthermore, these advancements have been successfully adapted to MLLMs for video understanding and reasoning~\cite{feng2025video,li2025videochat,chen2025exploring,geminithinking}. This progress naturally raises a question: can these models perform complex video reasoning in a manner comparable to human experts? 

However, existing evaluation benchmarks for video reasoning~\cite{yang2024thinking,he2024mmworld,zhao2025mmvu,qi2025vcr,hu2025video,li2024mvbench,liu2024tempcompass,cheng2025v} are limited by their predominant focus on assessing the visual perception and grounding capabilities of models, where questions that can answered based on explicit prompts or isolated visual cues (e.g., ``What is the woman wearing?''). Such benchmarks do not fully capture the intricacies of real-world reasoning, where humans must actively search for, integrate, and analyze multiple clues before reaching a conclusion. Empirical results show that even models with advanced thinking abilities~\cite{geminithinking} achieve only marginal gains (e.g., from 68.3\% to 69.4\%) on these benchmarks~\cite{fu2024video}, which raises doubts about the extent to which these tasks require genuine reasoning.

To address this issue, we present Video-Holmes, a benchmark inspired by the reasoning process of Sherlock Holmes. It is designed to assess the complex video reasoning abilities of MLLMs and exam the factors contributing their correct and incorrect answers. As demonstrated in Table~\ref{tab: benchmark compare}, Video-Holmes differs from existing benchmarks in several key aspects: (1) We utilize suspense short films as the video sources with detailed manual annotations. These videos are characterized by rich elements of suspense, reasoning, and supernatural themes, making them particularly challenging for models to comprehend. (2) The questions in Video-Holmes require models to actively locate and connect multiple relevant visual clues scattered across different video segments to infer the final answer. As illustrated in Figure~\ref{fig: teasor}, models first need to identify the abnormal scene involving the man and then progressively integrate the extracted visual clues to deduce the cause of the man's death, just like the reasoning process of Sherlock Holmes. (3) We provide detailed analysis of models' reasoning processes, examining the factors that lead to both correct and incorrect answers. 
%\clearpage

Our comprehensive evaluation of state-of-the-art (SOTA) MLLMs reveals that, while these models generally excel at visual perception, they encounter substantial difficulties with integrating information and often miss critical clues. For example, the best-performing model, Gemini-2.5-Pro, achieves an accuracy of only 45\%, with most models scoring below 40\%.

We make the following contributions in this work:
\begin{itemize}
    \item We present Video-Holmes, a benchmark for complex video reasoning. Video-Holmes comprises 270 manually annotated suspense short films, along with 1,837 challenging questions, which require models to actively locate and link multiple relevant visual clues, offering the research community a high-quality and challenging video reasoning benchmark.
    \item We conduct extensive experiments on Video-Holmes to evaluate existing SOTA MLLMs. We analyze the reasoning processes of these models and observe that while they perform well in visual perception, they face significant challenges in integrating clues and frequently overlook critical clues. These observations provide valuable insights for future research.
\end{itemize}

\begin{table*}[!t]
  \caption{Comparison between \ourbench~and existing video reasoning benchmarks across several key aspects: the video source domain (\textbf{Domain}), annotation methodology (\textbf{Anno.}), the number of reasoning QA pairs (\textbf{RQA Pairs}), necessity for models to actively seek out clues (\textbf{Active Seeking}), necessity for models to link multiple clues (\textbf{Chain-of-Clues}), whether provide reasoning process analysis (\textbf{RPA}), and whether provide audio information (\textbf{Aud.}).}
  \label{tab: benchmark compare}
  \centering
\resizebox{1\textwidth}{!}{
\begin{tabular}{llccccccccc}
\toprule
Benchmarks  & Domain & Anno. & RQA Pairs & Active Seeking & Chain-of-Clues & RPA & Aud. \\
\midrule
VSI-Bench~\cite{yang2024thinking} & Indoor scenes &  A\&M   &   5,130  &  \ding{55} &  \ding{55} & \ding{55} &  \ding{55}       \\
MVBench~\cite{li2024mvbench}   &    Open    &  A   & 4,000   &  \ding{55} &  \ding{55} & \ding{55} &  \ding{55}   \\
TempCompass~\cite{liu2024tempcompass} & Open & A\&M & 7,540   &  \ding{55} &  \ding{55} & \ding{55} & \ding{55}   \\
Video-MMMU~\cite{hu2025video} & Academic &  M   &   900  &  \ding{55} &  \ding{55} & \ding{55} & $\checkmark$         \\
MMVU~\cite{zhao2025mmvu} & Science  &  M    &  3,000  &  \ding{55} &  \ding{55} & \ding{55} & $\checkmark$               \\
Video-MME~\cite{fu2024video} &  Open    &   M   &  1,944    &  \ding{55} &  \ding{55} & \ding{55} & $\checkmark$          \\
VCR-Bench~\cite{qi2025vcr} &    Open &  A\&M    & 1,034  &  \ding{55} &  \ding{55} & $\checkmark$  & \ding{55}   \\
\midrule
\textbf{\ourbench~(Ours)} &  Suspense short films    &   A\&M       &  1,837 & $\checkmark$ & $\checkmark$ & $\checkmark$ & $\checkmark$  \\
\bottomrule
\end{tabular}
}
\end{table*}

\section{Related Works}

{\bf MLLMs for Video Understanding and Reasoning.}  
The advancements in MLLMs for image understanding and reasoning~\cite{zhu2023minigpt,zheng2023minigpt,ge2023making,cao2024visdiahalbench} have enabled their extension to video-based tasks. Methods such as VideoChat~\cite{li2023videochat}, Video-ChatGPT~\cite{maaz2023video}, CogVLM2Video~\cite{hong2024cogvlm2}, InternVL~\cite{chen2024internvl,zhu2025internvl3}, LLaVA-Video~\cite{zhang2024video}, and Qwen-VL~\cite{wang2024qwen2,xu2025qwen2} treat videos as sequences of images, allowing large language models (LLMs) to perform video understanding and reasoning. CoT reasoning and reinforcement post-training strategies have been shown to enhance the reasoning capabilities of LLMs~\cite{o1,guo2025deepseek}, and recent work extends these techniques to MLLMs for multimodal reasoning tasks involving images~\cite{team2025kimi,liao2025improved,deng2025openvlthinker,shen2025vlm,xu2024llava,thawakar2025llamav} or videos~\cite{li2025videochat,feng2025video,zhang2025tinyllava,chen2025exploring,wang2025timezero}. For instance, Video-R1~\cite{feng2025video} and VideoChat-R1~\cite{li2025videochat} employ supervised fine-tuning (SFT) followed by reinforcement learning post-training using Group Relative Policy Optimization (GRPO)~\cite{shao2024deepseekmath} based on Qwen-VL-2.5~\cite{bai2025qwen2}. These methods achieve higher performance compared to Vanilla Qwen-VL-2.5 across diverse reasoning tasks.

{\bf Video Reasoning Benchmarks.}
Early video understanding benchmarks primarily assess model capabilities within specific scenarios. For instance, MSRVTT-QA~\cite{xu2017video}, ActivityNet-QA~\cite{yu2019activitynet}, and NExT-QA~\cite{xiao2021next} focus on fundamental tasks such as action recognition and video question answering. Recently, benchmarks like MMBench~\cite{xu2023mmbench}, TempCompass~\cite{liu2024tempcompass}, and MVBench~\cite{li2024mvbench} evaluate reasoning over short video clips, while LongVideoBench~\cite{wu2024longvideobench} and Video-MME~\cite{hu2025video} extend evaluations to longer video sequences. However, these tasks are generally straightforward and do not require complex reasoning. With the success of chain-of-thought (CoT) reasoning, there is increasing interest in advancing video reasoning in more challenging scenarios. Benchmarks such as MMVU~\cite{zhao2025mmvu} and VideoMMMU~\cite{he2024mmworld} evaluate reasoning in academic and scientific domains, while VSI-Bench~\cite{yang2024thinking} focuses on indoor environments. The recent VCR-Bench~\cite{qi2025vcr} introduces a benchmark specifically designed to assess CoT reasoning in video tasks. Despite these developments, such benchmarks primarily evaluate visual perception and grounding abilities, with questions that can be answered based on explicit prompts or isolated visual cues and do not fully capture the intricacies of real-world reasoning. In contrast, Video-Holmes  require models to actively locate and connect multiple relevant visual clues, engaging them in a more complex and demanding reasoning scenario.

\section{\ourbench}

\begin{figure*}[!t]
	\centering
	\includegraphics[width=\textwidth]{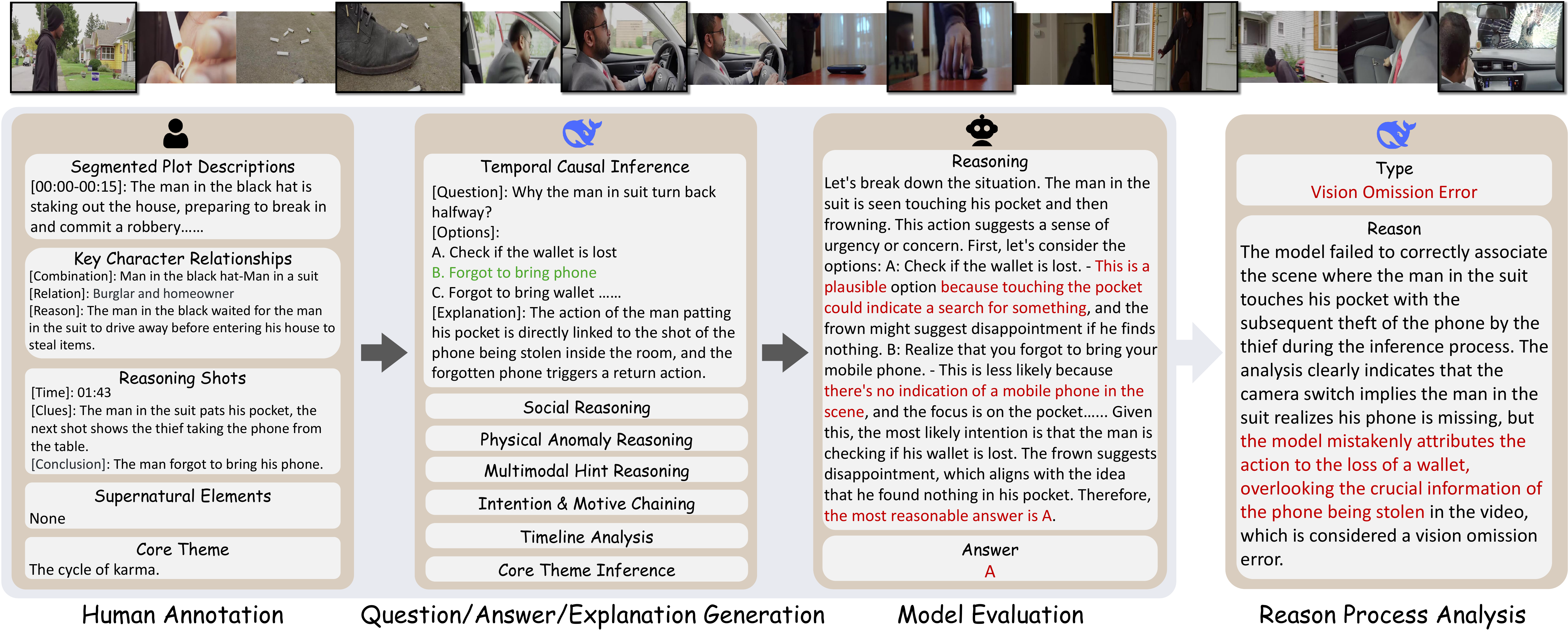}
\caption{Construction and evaluation pipeline of Video-Holmes. We select 270 high-quality suspense short films for human annotation. Next, we design 7 challenging tasks and employ DeepSeek to generate questions. Finally, we evaluate SOTA MLLMs and use DeepSeek to analyze their responses.}
\label{fig: pipeline}
\vspace{-5pt}
\end{figure*}

As shown in Figure~\ref{fig: pipeline}, the construction of Video-Holmes involves three steps: video collection and annotation, task definition, and question-answer-explanation generation.

\textbf{Video Collection and Annotation.} Suspense short films serve as an ideal source for evaluating the complex video reasoning capabilities of MLLMs, as they are characterized by compact narratives enriched with hints, plot twists, and supernatural elements. We utilize the keyword ``suspense short films'' to search videos from YouTube with durations between 1 and 5 minutes. We incorporate nine subkeywords\footnote{Details are provided in Appendix~\ref{app C}.} in this process to ensure diversity. From the initial pool of over 2,500 videos with audio information retrieved through our search, we manually curated a subset of 270 high-quality, reasoning-reach short films with a rigorous annotation process. Each film is annotated following a structured template that considers the following aspects: 
\begin{itemize}
    \item \textbf{Segmented Plot Descriptions:} Annotators are asked to divide the video into segments based on the progression of the storyline, and provide detailed descriptions for each segment.
    \item \textbf{Key Character Relationships:} Annotators are asked to present the relationships between key characters in the video, along with evidence that supports the identification.
    \item \textbf{Reasoning Shots:} Annotators are asked to identify reasoning shots in the video, providing timestamps, visual clues, and the inferred conclusions associated with these shots.
    \item \textbf{Supernatural Elements:} Annotators are asked to specify any supernatural elements present in the videos and the implications they introduce, whether positive or negative.
    \item \textbf{Core Theme:} Annotators are asked to summarize the core themes of the videos.
\end{itemize}

These diverse short films with intricate reasoning chains, along with high-quality and well-formatted manual annotations ensures the reliability and quality of Video-Holmes.

\begin{figure*}[!t]
	\centering
	\includegraphics[width=0.95\textwidth]{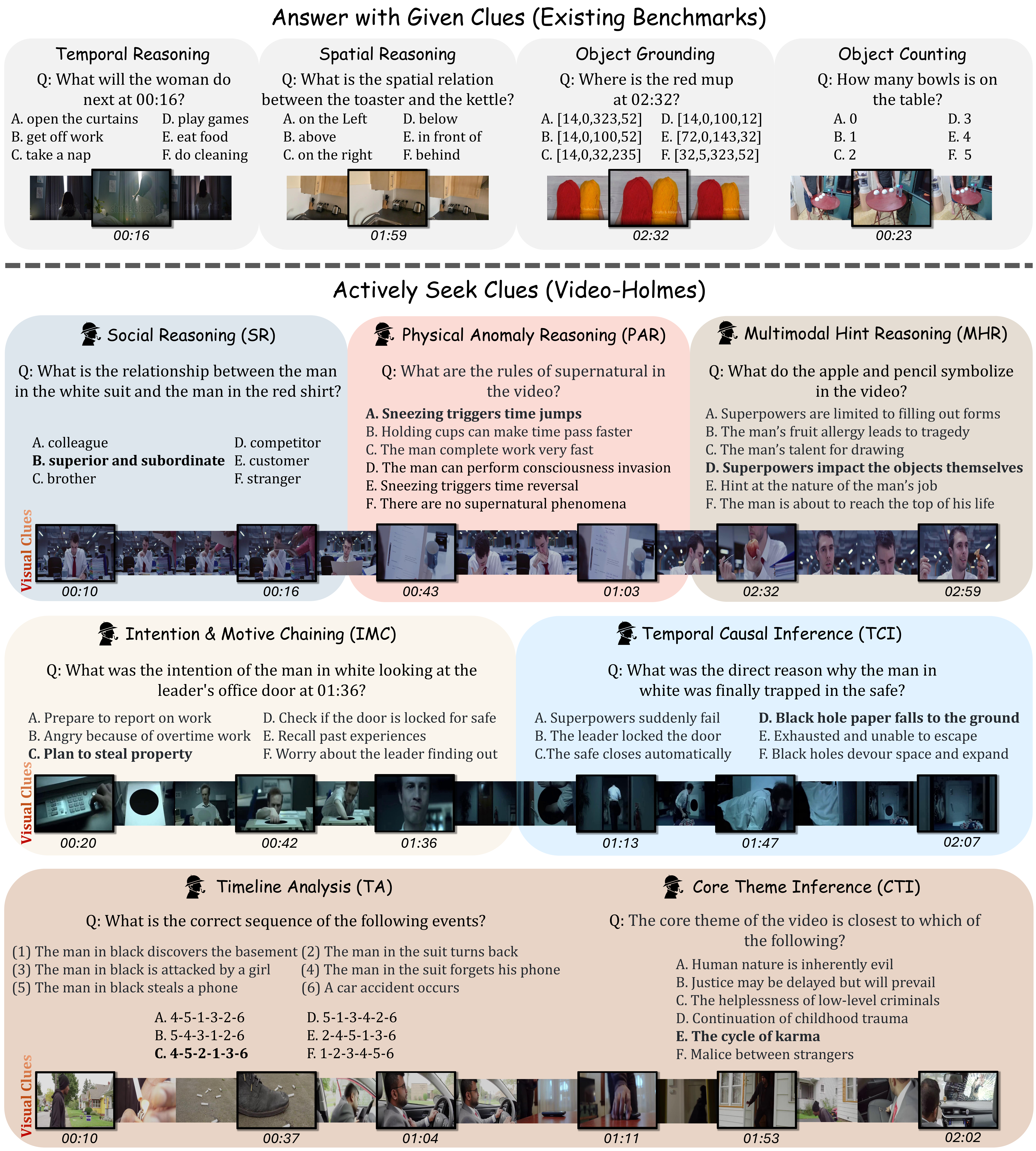}
	\caption{Comparison of question types between Video-Holmes and existing benchmarks. Existing benchmarks primarily involve clue-given questions, where models depend on explicitly provided clues to derive answers. In contrast, Video-Holmes adopts an active seeking paradigm, requiring models to actively locate and connect multiple relevant visual clues scattered across different video segments. (Key frames are marked with black boxes and magnified.)}
    \vspace{-15pt}
\label{fig: task definition}

\end{figure*}

\textbf{Task Definition.} To comprehensively evaluate the differences in MLLMs’ capabilities for complex video reasoning from multiple perspectives, we define seven distinct reasoning tasks for Video-Holmes. As illustrated in Figure~\ref{fig: task definition}, different from existing benchmarks that primarily designed around clue-given questions, Video-Holmes focus on tasks that require models to actively locate and connect multiple relevant visual clues scattered across different video segments:
\begin{itemize}
    \item \textbf{Social Reasoning (SR):} Inferring social relationships between characters. This includes identifying identity associations across time (e.g., the same man in youth and old age).
    \item \textbf{Physical Anomaly Reasoning (PAR):} Identifying scenes in the video that deviate from real-world norms and reasoning about their underlying rules or implicit meanings.
    \item \textbf{Multimodal Hint Reasoning (MHR):} Decoding cues or fact from multimodal hints, such as semantic implications of camera movements or gradual changes in object positions.
    \item \textbf{Intention \& Motive Chaining (IMC):} Observing characters' actions or environmental cues to disentangle surface behaviors from underlying behavioral intentions.
    \item \textbf{Temporal Causal Inference (TCI):} Inferring causal mechanisms between events across time and space using cinematic language and multimodal clues.
    \item \textbf{Timeline Analysis (TA):} Integrating and reconstructing the narrative storyline of the film.
    \item \textbf{Core Theme Inference (CTI):} Extracting the core theme or deeper meaning of the video by analyzing its plot, dialogues, and symbolic elements.
\end{itemize}

\textbf{Question-Answer Generation.} We utilize DeepSeek-R1~\cite{guo2025deepseek} with advanced reasoning capabilities to automatically generate questions based on formatted manual annotations and predefined question types. Each question is generated by strictly adhering to the provided information, with manual sampling inspection to ensure quality and relevance. Additionally, the model is required to provide correct answer explanations for each question, which are used to compare and analyze the model's reasoning process. Please refer to Appendix~\ref{app B} for details.

After data verification and annotation, we have ultimately constructed a dataset comprising 270 videos and 1,837 question-answer pairs. The key statistics of Video-Holmes are presented in Appendix~\ref{app C}.

\section{Experiments}

\subsection{Setup}

{\bf Evaluation Models.} We conduct an evaluation of several mainstream MLLMs, including the open-source models: InternVL2.5 (8B)~\cite{chen2024expanding}, InternVL3 (8B)~\cite{zhu2025internvl3}, Qwen2.5-VL (7B, 32B)~\cite{bai2025qwen2}, and Qwen2.5-Omni (7B)~\cite{xu2025qwen2}. Additionally, we assess open-source models that incorporate RL post-training based on Qwen2.5-VL (7B): SEED-Bench-R1~\cite{chen2025exploring}, Video-R1~\cite{feng2025video}, and VideoChat-R1~\cite{li2025videochat}. We also include several advanced closed-source models in our evaluation: Gemini-2.0-Flash~\cite{pichai2024introducing}, Gemini-2.0-Flash-Thinking~\cite{geminithinking}, Gemini-1.5-Pro~\cite{gemini2}, Gemini-2.5-Pro~\cite{gemini25pro}, GPT-4o~\cite{4o}, OpenAI o4-mini~\cite{o4mini}, Claud 3.5 Sonnet~\cite{claud}, and Claud 3.7 Sonnet~\cite{claud}.

{\bf Implementation Details.} For models with native video input support, such as Qwen-VL and Gemini, videos were processed directly without additional pre-processing. For models lacking native video input capabilities (e.g., GPT-4o), frames were uniformly extracted from the video along with corresponding timestamp annotations, and multi-image input was utilized for evaluation. To ensure a fair comparison, all models were deployed following their official guidelines and using the officially released checkpoints. During inference, models were required to first generate a reasoning process before providing the final answer. Specifically, the models were instructed to produce a step-by-step solution to the given question. For further details regarding model implementation and evaluation prompts, please refer to Appendix~\ref{app A} and~\ref{app B}.

\definecolor{mygray}{HTML}{E6F0E8}
\begin{table*}[!t]
  \caption{Results of various models on Video-Holmes, where \textbf{SR} stands for Social Reasoning; \textbf{IMC} stands for Intention \& Motive Chaining; \textbf{TCI} stands for Temporal Causal Inference; \textbf{TA} Timeline Analysis; \textbf{MHR} stands for Multimodal Hint Reasoning; \textbf{PAR} stands for Physical Anomaly Reasoning; \textbf{CTI} stands for Core Theme Inference. \colorbox{color1}{Blue} represents the vanilla model, while \colorbox{color3}{Green} represents its corresponding thinking version with RL post-training.}
  \label{tab: main results}
  \centering
\resizebox{1\textwidth}{!}{
\begin{tabular}{lccccccccc}
\toprule
Model &Frames &SR  &IMC  &TCI  &TA  &MHR  &PAR &CTI  &Overall \\
\midrule \midrule
\vspace{-4mm} \\
\multicolumn{10}{>{\columncolor{Gray}}c}{\textit{Open Source Models}} \\
\vspace{-2.8mm} \\
InternVL2.5-8B &32 & 28.0 &32.2 &21.5 & 07.7  &25.7 &23.8 &22.6 &23.8 \\
InternVL3-8B  &32 & 29.5 & 40.7 & 37.9 & 35.1 & 24.6 & 38.9 & 24.1 & 32.3 \\
Qwen2.5-Omni-7B  & 32 & 27.1 & 19.9 & 13.9 & 7.5  & 14.8 & 14.9 & 13.7 & 16.4 \\
Qwen2.5-VL-32B  &32 & 43.2 & 44.2 & 31.5 & 51.0   & 36.4 & 31.4 & 32.2 & 38.4 \\
\cellcolor{color1}Qwen2.5-VL-7B &\cellcolor{color1}32 &\cellcolor{color1}38.4 &\cellcolor{color1}34.8 &\cellcolor{color1}17.6 &\cellcolor{color1}30.0   &\cellcolor{color1}27.1 &\cellcolor{color1}18.6 &\cellcolor{color1}25.2 &\cellcolor{color1}27.8 \\
\cellcolor{color3}SEED-Bench-R1 &\cellcolor{color3}32 &\cellcolor{color3}42.8 &\cellcolor{color3}35.1 &\cellcolor{color3}25.6 &\cellcolor{color3}40.5 &\cellcolor{color3}29.2 &\cellcolor{color3}29.9 &\cellcolor{color3}32.6 &\cellcolor{color3}33.5 \\
\cellcolor{color3}VideoChat-R1 &\cellcolor{color3}32 &\cellcolor{color3}42.1 &\cellcolor{color3}38.8 &\cellcolor{color3}24.5 &\cellcolor{color3}39.5 &\cellcolor{color3}29.5 &\cellcolor{color3}27.8 &\cellcolor{color3}29.3 &\cellcolor{color3}33.0   \\
\cellcolor{color3}Video-R1 &\cellcolor{color3}32 &\cellcolor{color3}48.6 &\cellcolor{color3}41.7 &\cellcolor{color3}28.9 &\cellcolor{color3}34.5 &\cellcolor{color3}31.0   &\cellcolor{color3}33.5 &\cellcolor{color3}35.9 &\cellcolor{color3}36.5 \\
\vspace{-2.8mm} \\
\midrule
\multicolumn{10}{>{\columncolor{Gray}}c}{\textit{Closed Source Models}} \\
\vspace{-2.8mm} \\
GPT-4o & 32 & 50.0   & 49.6 & 38.8 & 30.0   & 44.0   & 39.2 & 37.0   & 42.0   \\
OpenAI o4-mini &32 & 36.3 & 31.2 & 20.5 & 34.0   & 30.1 & 30.9 & 27.4 & 29.9 \\
Claud 3.5 Sonnet & 20 & 48.6 & 43.5 & 30.8 & 41.0   & 39.8 & 36.6 & 33.7 & 39.3 \\ 
Claud 3.7 Sonnet & 20 & 45.9 & 48.2 & 33.7 & 39.5 & 40.7 & 39.7 & 38.1 & 41.0   \\
Gemini-1.5-Pro &-  & 52.1 & 48.2 & 34.4 & 26.0   & 39.2 & 46.4 & 38.9 & 41.2 \\
Gemini-2.5-Pro &-  & 46.6 & 49.3 & 46.9 & 53.0   & 40.1 & 44.3 & 37.4 & 45.0   \\
\cellcolor{color1}Gemini-2.0-Flash  &\cellcolor{color1}- &\cellcolor{color1}41.8 &\cellcolor{color1}33.7 &\cellcolor{color1}23.1 &\cellcolor{color1}20.5 &\cellcolor{color1}30.1 &\cellcolor{color1}26.8 &\cellcolor{color1}33.7 &\cellcolor{color1}30.6 \\
\cellcolor{color3}Gemini-2.0-Flash-Thinking &\cellcolor{color3}- &\cellcolor{color3}43.4 &\cellcolor{color3}46.9 &\cellcolor{color3}43.1 &\cellcolor{color3}51.0   &\cellcolor{color3}37.9 &\cellcolor{color3}43.6 &\cellcolor{color3}39.3 &\cellcolor{color3}43.1 \\
\bottomrule
\end{tabular}
}
\end{table*}
\begin{table*}[!t]
\centering
\caption{Thinking model performances on Video-Holmes and other benchmarks.}
\vspace{-0.6em}
\resizebox{\linewidth}{!}{
\begin{tabular}{lccccc}
\toprule
Model &MMVU &Video-MME & Video-MMMU & MVBench & \textbf{Video-Holmes} \\
\midrule
Gemini-2.0-Flash &65.2  &68.3 & 54.3 & 55.3 & 30.6 \\
Gemini-2.0-Flash-Thinking &64.5 \small \textcolor{deepred}{(-0.7)} &69.4 \small \textcolor{mydeepgreen}{(+0.9)} & 56.4 \small \textcolor{mydeepgreen}{(+1.9)}& 57.4 \small \textcolor{mydeepgreen}{(+2.1)} & 43.1 \small \textbf{\textcolor{mydeepgreen}{(+12.5)}} \\ 
\bottomrule
\end{tabular}}
\vspace{-1em}
\label{tab: other bench}
\end{table*}

\subsection{Main Results}

Table~\ref{tab: main results} presents the performance of each models on Video-Holmes benchmark. Most models achieve an accuracy below 40\%, with the best-performing model, Gemini-2.5-Pro, reaching an overall accuracy of 45\%. The widely-used open-source models, Qwen2.5-VL (7B) achieves an overall accuracy of 27.8\%, far worse than its performance on other video reasoning benchmarks. This performance gap suggests that the Video-Holmes benchmark introduces unique challenges that are particularly demanding for current MLLMs in video reasoning tasks. 

Models trained with thinking strategies exhibit notable improvements over their vanilla version. For instance, Gemini-2.0-Flash-Thinking demonstrates a 12.5\% performance gain compared to Gemini-2.0-Flash. This observation indicates that the Video-Holmes benchmark imposes substantial reasoning challenges and effectively distinguishes models' reasoning abilities. In contrast, other benchmarks do not reflect this pattern, as illustrated in Table~\ref{tab: other bench}.

Model performance across seven reasoning tasks in Video-Holmes remains relatively even, with most models achieving accuracy below 40\% for each task. This highlights that each task in Video-Holmes poses substantial challenges, requiring advanced reasoning capabilities from existing methods.

\subsection{Analytical Study} 

\begin{table*}[!t]
\caption{Reasoning process analysis results. Where \textbf{VPE} represents visual perception error, \textbf{VOE} represents visual omission error, \textbf{RE} represents reasoning error, \textbf{TRAW} represents think right answer wrong, \textbf{TWAR} represents think wrong answer right, and \textbf{TRAR} represents think right answer right.}
\label{tab: thinking}
\centering
\resizebox{\textwidth}{!}{
\begin{tabular}{l|cccccccc|cccc}
\toprule
\multicolumn{1}{c|}{\multirow{3}{*}{Model}} &\multicolumn{8}{c|}{Answer Wrong} &\multicolumn{4}{c}{Answer Right} \\
\multicolumn{1}{c|}{} &\multicolumn{2}{c}{VPE} &\multicolumn{2}{c}{VOE}  &\multicolumn{2}{c}{RE}  &\multicolumn{2}{c|}{TRAW}  &\multicolumn{2}{c}{TRAR}  &\multicolumn{2}{c}{TWAR} \\

\cmidrule(lr){2-3} \cmidrule(lr){4-5} \cmidrule(lr){6-7} \cmidrule(lr){8-9} \cmidrule(lr){10-11} \cmidrule(lr){12-13} 

\multicolumn{1}{c}{}  &\multicolumn{1}{|l}{Count} &\multicolumn{1}{l}{Ratio} &\multicolumn{1}{l}{Count} &\multicolumn{1}{l}{Ratio} &\multicolumn{1}{l}{Count} &\multicolumn{1}{l}{Ratio} &\multicolumn{1}{l}{Count} &\multicolumn{1}{l|}{Ratio} &\multicolumn{1}{l}{Count} &\multicolumn{1}{l}{Ratio} &\multicolumn{1}{l}{Count} &\multicolumn{1}{l}{Ratio} \\

\midrule \midrule
\multicolumn{13}{>{\columncolor{Gray}}c}{\textit{Open Source Models}} \\
InternVL2.5-8B            & 66        & 0.06      & 331       & 0.28      & 740      & 0.63     & 32         & 0.03       & 332        & 0.79       & 89         & 0.21       \\
InternVL3-8B              & 32        & 0.03      & 248       & 0.22      & 810      & 0.73     & 22         & 0.02       & 419        & 0.76       & 135        & 0.24       \\
Qwen2.5-Omni-7B           & 26        & 0.02      & 398       & 0.36      & 643      & 0.59     & 29         & 0.03       & 236        & 0.82       & 53         & 0.18       \\
Qwen2.5-VL-32B            & 44        & 0.04      & 430       & 0.39      & 610      & 0.56     & 6          & 0.01       & 363        & 0.78       & 103        & 0.22       \\

\cellcolor{color1}Qwen2.5-VL-7B &\cellcolor{color1}15        &\cellcolor{color1}0.02      &\cellcolor{color1}209       &\cellcolor{color1}0.27      &\cellcolor{color1}544      &\cellcolor{color1}0.70      &\cellcolor{color1}5          &\cellcolor{color1}0.01       &\cellcolor{color1}457        &\cellcolor{color1}0.87       &\cellcolor{color1}67         &\cellcolor{color1}0.13       \\
\cellcolor{color3}SEED-Bench-R1             &\cellcolor{color3}32        &\cellcolor{color3}0.03      &\cellcolor{color3}325       &\cellcolor{color3}0.30       &\cellcolor{color3}688      &\cellcolor{color3}0.63     &\cellcolor{color3}54         &\cellcolor{color3}0.05       &\cellcolor{color3}543        &\cellcolor{color3}0.80        &\cellcolor{color3}133        &\cellcolor{color3}0.20        \\
\cellcolor{color3}VideoChat-R1              &\cellcolor{color3}33        &\cellcolor{color3}0.03      &\cellcolor{color3}306       &\cellcolor{color3}0.26      &\cellcolor{color3}844      &\cellcolor{color3}0.71     &\cellcolor{color3}12         &\cellcolor{color3}0.01       &\cellcolor{color3}494        &\cellcolor{color3}0.85       &\cellcolor{color3}88         &\cellcolor{color3}0.15       \\
\cellcolor{color3}Video-R1                  &\cellcolor{color3}46        &\cellcolor{color3}0.04      &\cellcolor{color3}415       &\cellcolor{color3}0.36      &\cellcolor{color3}681      &\cellcolor{color3}0.59     &\cellcolor{color3}15         &\cellcolor{color3}0.01       &\cellcolor{color3}504        &\cellcolor{color3}0.83       &\cellcolor{color3}101        &\cellcolor{color3}0.17       \\

\midrule
\multicolumn{13}{>{\columncolor{Gray}}c}{\textit{Closed Source Models}} \\
GPT-4o                    & 17        & 0.02      & 250       & 0.24      & 745      & 0.73     & 13         & 0.01       & 720        & 0.94       & 43         & 0.06       \\
o4-mini                   & 36        & 0.03      & 354       & 0.29      & 840      & 0.68     & 11         & 0.01       & 513        & 0.94       & 34         & 0.06       \\
Claud 3.5 Sonnet          & 18        & 0.02      & 318       & 0.31      & 695      & 0.67     & 11         & 0.01       & 687        & 0.92       & 59         & 0.08       \\
Claud 3.7 Sonnet          & 9         & 0.01      & 470       & 0.44      & 584      & 0.54     & 10         & 0.01       & 612        & 0.86       & 103        & 0.14       \\
Gemini-1.5-Pro            & 75        & 0.07      & 250       & 0.24      & 692      & 0.67     & 18         & 0.02       & 637        & 0.85       & 114        & 0.15       \\
Gemini-2.5-Pro            & 31        & 0.03      & 210       & 0.22      & 718      & 0.75     & 4          & 0.01       & 725        & 0.88       & 98         & 0.12       \\
\cellcolor{color1}Gemini-2.0-Flash          &\cellcolor{color1}33        &\cellcolor{color1}0.03      &\cellcolor{color1}242       &\cellcolor{color1}0.20       &\cellcolor{color1}893      &\cellcolor{color1}0.76     &\cellcolor{color1}13         &\cellcolor{color1}0.01       &\cellcolor{color1}542        &\cellcolor{color1}0.89       &\cellcolor{color1}65         &\cellcolor{color1}0.11       \\
\cellcolor{color3}Gemini-2.0-Flash-Thinking &\cellcolor{color3}31        &\cellcolor{color3}0.03      &\cellcolor{color3}210       &\cellcolor{color3}0.22      &\cellcolor{color3}718      &\cellcolor{color3}0.75     &\cellcolor{color3}4          &\cellcolor{color3}0.01       &\cellcolor{color3}774        &\cellcolor{color3}0.94       &\cellcolor{color3}49         &\cellcolor{color3}0.06      \\
\bottomrule
\end{tabular}
}
\end{table*}

\begin{figure*}[!t]
	\centering
	\includegraphics[width=\textwidth]{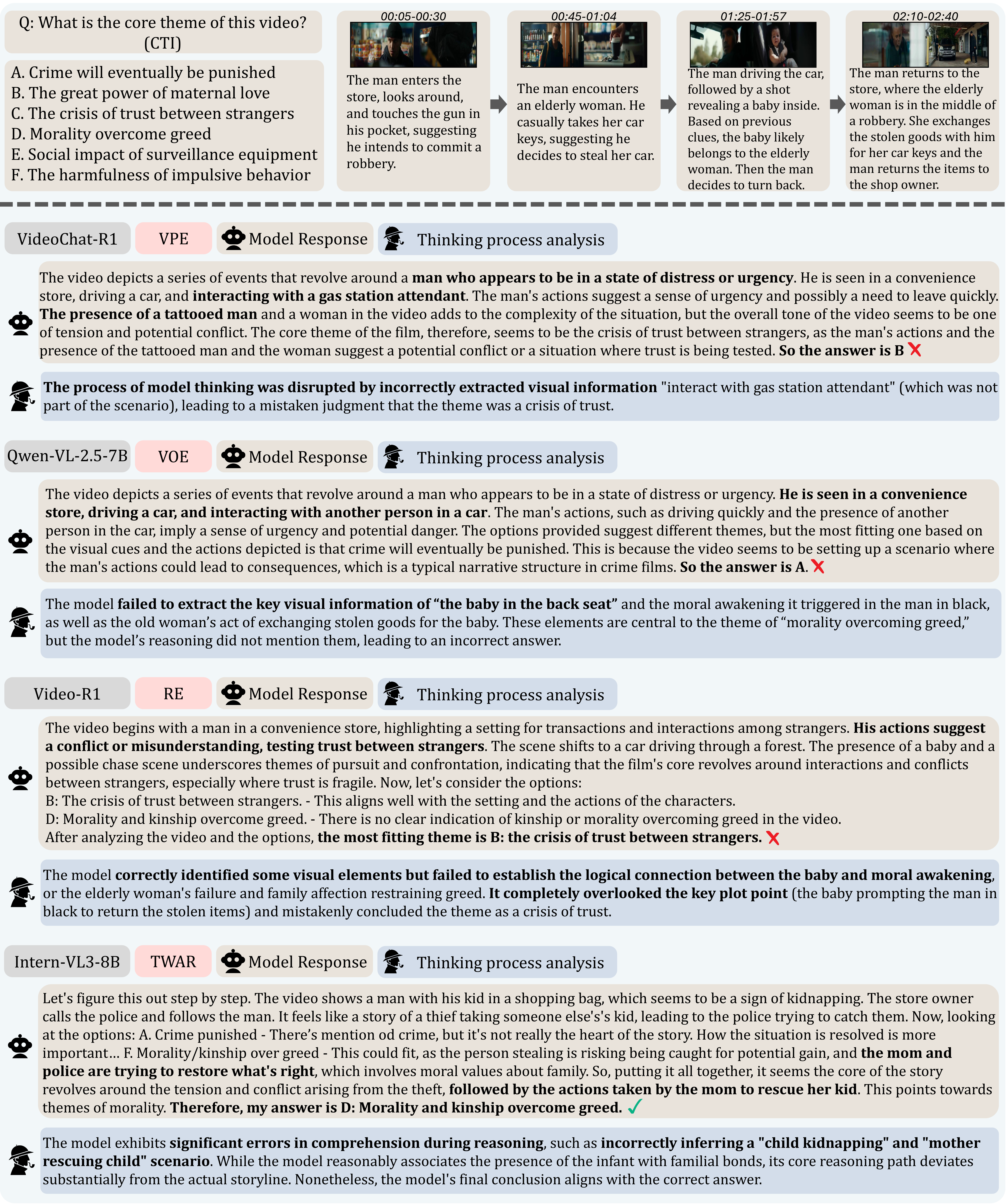}
\caption{Example of model reasoning processes on Video-Holmes. VideoChat-R1 misinterprets visual information, incorrectly perceiving a tattooed man, Qwen2.5-VL overlooks critical visual clues (the baby), Video-R1 fails to establish logical connections between the visual clues, and Intern-VL3-8B guesses the right option with a wrong reasoning process.}
\vspace{-1.7em}
\label{fig: case study}
\end{figure*}

\textbf{Reasoning Process Analysis.}
We analyze the factors contributing to the model's answers by comparing its reasoning process with human-annotated descriptions and answer explanations. Specifically, we categorize the main causes of incorrect answers into the following four types:
\begin{itemize}
    \item \textbf{Visual Perception Error (VPE):} The model extracts incorrect visual information for analysis, leading to an incorrect answer.
    \item \textbf{Visual Omission Error (VOE):} The model omits critical visual information (i.e., key objects or events), resulting in an incorrect answer. 
    \item \textbf{Reasoning Error (RE):} The model makes errors during the reasoning process, such as misinterpreting or incorrectly associating multiple visual clues.
    \item \textbf{Think Right Answer Wrong (TRAW):} The model's reasoning is largely aligned with the ground-truth explanation, but it selects an incorrect option when providing the final answer.
\end{itemize}
For correctly answered questions, we define the following two categories: 
\begin{itemize}
    \item \textbf{Think Wrong Answer Right (TWAR):} The model's reasoning process deviates significantly from the ground-truth explanation, yet it arrives at the correct answer. 
    \item \textbf{Think Right Answer Right (TRAR):} The model's reasoning process is largely aligned with the ground-truth explanation and produces answers consistent with its reasoning.
\end{itemize}

We provide the human annotations, questions, the model's reasoning outputs (if validated), and the type definitions as inputs to DeepSeek-R1~\cite{guo2025deepseek}, prompting\footnote{Detailed in Appendix~\ref{app B}} it to perform the analysis.

The results in Table~\ref{tab: thinking} and Figure~\ref{fig: case study} show that both open-source and closed-source models generally demonstrate the ability to accurately extract visual information and provide answers consistent with their reasoning processes. Approximately 35\% of errors are attributed to the omission of critical visual information, while a larger proportion (around 60\%) stems from challenges in logical comprehension of multiple visual clues (Reasoning Errors). 

For correctly answered questions, the proportion of responses based on valid reasoning (TRAR) exceeds 80\% across most models. This highlights the difficulty of the Video-Holmes benchmark, where models struggle to infer correct answers through inconsistent reasoning.

\textbf{Number of Input Frames.} We analyze the performance of several models using different numbers of input frames. The results in Table~\ref{tab: anaresults} (a) indicate that increasing the number of input frames generally improves model performance, but does not lead to substantial gains. This observation indicates that in most cases, the visual information provided is sufficient, and the key challenge lies in the model's ability to integrate and interpret visual clues effectively.

\textbf{Audio Input.} We evaluate the performance of several models with audio input as an additional modality. Table~\ref{tab: anaresults} (b) demonstrates that integrating audio input enhances model performance, especially in social reasoning tasks where conversational cues offer essential insights into interpersonal dynamics. These results underscore the importance of audio information in multimodal reasoning.

\textbf{Reasoning or Not.} We conduct experiments where models directly generate answers without using CoT prompts. The results in Table~\ref{tab: anaresults} (c) demonstrate that for stronger closed-source models, CoT prompting leads to higher accuracy compared to directly answering. In contrast, weaker open-source models exhibit the opposite trend. This suggests that the effectiveness of reasoning is contingent on the model's overall capability—only models with sufficiently strong reasoning abilities can fully benefit from CoT prompts. Conversely, weaker models may amplify errors during CoT reasoning.

\textbf{Text-only Input.} We conduct experiments using three types of text-only inputs for advanced reasoning models~\cite{guo2025deepseek,o3}: (1) human-annotated movie plots and key clues (excluding reasoning conclusions), (2) frame-level captions generated by Qwen2.5-VL (sampled at one frame per second), and (3) video-level captions generated by Gemini-2.5-pro. Table~\ref{tab: anaresults} (d) shows that models achieve around 90\% accuracy with human annotations, while performance drops significantly with frame-level and video-level captions. This can be attributed to human annotations, which provide logical connections between critical visual clues in the video. In contrast, video-level and frame-level captions may overlook key visual details or offer incorrect logical interpretations, leading to reasoning errors. These findings underscore the challenge posed by Video-Holmes in requiring models to locate and and capture logical relationships within multiple visual clues.

\begin{table*}[!t]
  \caption{Analyze experiment results on Video-Holmes. \textbf{HA} stands for human annotation; \textbf{FLC} stands for frame-level caption; \textbf{VLC} stands for video-level caption.}
  \label{tab: anaresults}
  \centering
  \begin{tabular}{cc}
    \begin{minipage}{.365\textwidth}
      %\captionsetup{font=scriptsize}
      \centering
      \subcaption{ Number of input frames}
      \vspace{-2mm}
      \resizebox{\linewidth}{!}{
      \begin{tabular}{lcc}
      \toprule
      Model &Frames &Acc \\
      \midrule \midrule
      Qwen2.5-VL-7B  &64 &30.2 \textcolor{mydeepgreen}{(+2.4)} \\
      Qwen2.5-VL-7B  &80 & 33.0 \textcolor{mydeepgreen}{(+5.2)}\\
      Video-R1  &64  &37.4 \textcolor{mydeepgreen}{(+1.2)}\\
      Video-R1  &80 & 38.5 \textcolor{mydeepgreen}{(+2.0)}\\
      \midrule
      GPT-4o  &40 & 43.3 \textcolor{mydeepgreen}{(+1.3)} \\
      GPT-4o  &50 & 44.6 \textcolor{mydeepgreen}{(+2.6)} \\
      \bottomrule
      \end{tabular}
      }
    \end{minipage}%
    &
    \begin{minipage}{.398\textwidth}
      %\captionsetup{font=scriptsize}
      \centering
      \subcaption{ Audio input}
      \vspace{-2mm}
      \resizebox{\linewidth}{!}{
      \begin{tabular}{lccc}
      \toprule
      Model & Audio & SR & Overall \\
      \midrule \midrule
      Qwen2.5-Omni-7B &\ding{55} & 27.1 & 16.4 \\
      Qwen2.5-Omni-7B  &$\checkmark$ & 38.4 & 24.4 \\
      \midrule
      Gemini-2.5-Pro  &\ding{55} & 46.6 & 45.0 \\
      Gemini-2.5-Pro  &$\checkmark$ & 54.8 & 51.3 \\
      Gemini-1.5-Pro &\ding{55} & 52.1 & 41.2 \\
      Gemini-1.5-Pro &$\checkmark$ & 59.6 & 45.7 \\
      \bottomrule
      \end{tabular}
      }
    \end{minipage}%
    \\ \\
    \begin{minipage}{.4\textwidth}
      %\captionsetup{font=scriptsize}
      \centering
      \subcaption{Reasoning or not}
      \vspace{-2mm}
      \resizebox{\linewidth}{!}{
      \begin{tabular}{lccc}
      \toprule
      Model & Frames& Reasoning &Acc \\
      \midrule \midrule
      Qwen2.5-VL-7B &32  &$\checkmark$ & 27.8 \\
      Qwen2.5-VL-7B &32 &\ding{55} & 29.4 \\
      Video-R1 &32  &$\checkmark$ & 36.5 \\
      Video-R1 &32  &\ding{55}  &  28.2 \\
      \midrule
      Gemini-2.0-Flash &- &$\checkmark$ & 30.6\\
      Gemini-2.0-Flash &- &\ding{55} & 28.5 \\
      \bottomrule
      \end{tabular}
      }
    \end{minipage}%
    & 
\begin{minipage}{.394\textwidth}
      %\captionsetup{font=scriptsize}
      \centering
      \subcaption{ Text-only input}
      \vspace{-2mm}
      \resizebox{\linewidth}{!}{
      \begin{tabular}{lcc}
      \toprule
      Model &Input &Acc \\
      \midrule \midrule
      DeepSeek-R1 \ \ \ \ \ \ \ \ & \ \ \ \ \ \ \ \ \ FLC  \ \ \ \ \ \ \ \ \ & 31.2 \\
      DeepSeek-R1  \ \ \ \ \ \ \ \ & \ \ \ \ \ \ \ \ \ VLC \ \ \ \ \ \ \ \  \ & 64.6\\
      DeepSeek-R1   \ \ \ \ \ \ \ \ & \ \ \ \ \ \ \ \ \ HA \ \ \ \ \ \ \ \ \ & 92.0  \\
      OpenAI o3  \ \ \ \ \ \ \ \ & \ \ \ \ \ \ \ \ \ FLC \ \ \ \ \ \ \ \ \ & 25.4 \\
      OpenAI o3  \ \ \ \ \ \ \ \ &  \ \ \ \ \ \ \ \ \ VLC \ \ \ \ \ \ \ \ \ & 61.3  \\
      OpenAI o3 \ \ \ \ \ \ \ \ &  \ \ \ \ \ \ \ \ \ HA \ \ \ \ \ \ \ \ \ & 89.7 \\
      \bottomrule
      \end{tabular}
      }
    \end{minipage}
  \end{tabular}
\end{table*}

\section{Conclusion and Discussion}

In this work, we propose Video-Holmes, a benchmark designed to evaluate the complex video reasoning capabilities of MLLMs. Video-Holmes consists of 1,837 questions derived from 270 manually annotated suspense short films, which spans seven carefully designed tasks that require models to actively locate and connect multiple relevant visual clues scattered across different video segments. We conduct a detailed analysis of model reasoning processes, examining the factors that lead to both correct and incorrect answers. Our comprehensive evaluation of state-of-the-art MLLMs reveals that, while these models generally excel at visual perception, they encounter substantial difficulties with integrating information and often miss critical clues. We aim that Video-Holmes can serve as a \textit{``Holmes-test''} for multimodal reasoning, motivating models to reason more like humans and emphasizing the ongoing challenges in this field.

\clearpage

\appendix

\section*{Appendix}

\section{Model Implementation Details}
\label{app A}

\textbf{QwenVL:} We utilize the official checkpoints for different QwenVL models: \texttt{Qwen/Qwen2.5-VL-7B-Instruct} for QwenVL-2.5-7B, \texttt{Qwen/Qwen2.5-VL-7B-Instruct} for QwenVL-2.5-7B, \texttt{Qwen/Qwen2.5-VL-32B-Instruct} for QwenVL-2.5-32B, \texttt{Video-R1/Video-R1-7B} for Video-R1, and \texttt{OpenGVLab/VideoChat-R1-7B} for VideoChat-R1. The decoding configuration follows the settings provided in the official QwenVL-2.5 demo, with top-p set to 0.001 and temperature set to 0.01. During inference, we increase the frame resolution to 256 × 28 × 28 pixels to enhance visual fidelity.

\textbf{InternVL:} We utilize the official checkpoints for various InternVL models: \texttt{OpenGVLab/InternVL3-8B} for InternVL3-8B and \texttt{OpenGVLab/InternVL2-5-8B} for InternVL2.5-8B. The input image size is resized to $448 \times 448$ according to the official configuration.

\textbf{API Models:} We access the Gemini, GPT, and Claude model series via the official APIs provided by Google, OpenAI, and Anthropic. Specifically, for the GPT series, we use the official functions to retrieve image URLs and set the "detail" parameter to "low" as recommended.

\section{Prompt Template}
\label{app B}

\begin{tcolorbox}[colback=gray!5!white, colframe=gray!75!black, 
title=Question Generation Prompt, boxrule=0.5mm, width=\textwidth, arc=3mm, auto outer arc]
You are a professional logician and detective story enthusiast. I will provide you with the plot information of a short detective film. Your task is to create some reasoning multiple-choice questions for viewers who have watched this short detective film to test whether they truly understood it.

Requirements:\\
I will provide the following information:\\
(1) Segmental plot description: chronological plot description\\
(2) Key character relationships and cause positioning: the relationships between key characters in the film and the information that can be used to infer their relationships.\\
(3) Reasoning shots: key hint shots in the plot, plot twists, and climaxes.\\
(4) Supernatural elements: supernatural elements and rules appearing in the video.\\
(5) Main idea of the video: the possible main idea the video wants to express.\\

You need to create at least one multiple-choice question from each of the following seven types of reasoning (according to the specific requirements):\\

Type 1: Social Reasoning (SR)\\
Task description: infer the implicit social relationship network between characters through clothing style matching, group interaction topology analysis, including identity association across time spans (such as the same character in youth and old age).
Number of questions: Create one question for each combination based on the given key character relationships. If the [segmental plot description] provided includes the same character across time spans, create a question based on this.
Example 1: What is the relationship between the man at the beginning of the video and the skeleton at the end? (Answer: A. The same person)\\

Type 2: Intention and Motive Chaining (IMC)\\
Task description: infer the underlying psychological state and predict non-explicit behavioral intentions by observing characters' actions, expressions, or environmental clues. The question format must strictly follow the example: [Character] at [time] performed [very brief, non-key detail action description] with what intention/psychological state?
\end{tcolorbox}
\begin{tcolorbox}[colback=gray!5!white, colframe=gray!75!black, 
title=Question Generation Prompt (Continue), boxrule=0.5mm, width=\textwidth, arc=3mm, auto outer arc]
Number of questions: 1-2 questions
Example 1: What is the most likely psychological state of the man in black smoking at 00:03 in the video? (Answer: C. Anxiety about preparing for a crime)\\

Type 3: Temporal Causal Inference (TCI)\\
Task description: infer causal mechanisms between non-continuous temporal events through camera language and multimodal clues.
Number of questions: 1-2 questions
Example 1: Why did the man in the shirt die? (Answer: F. Overuse of superpowers)\\

Type 4: Timeline Analysis (TA)\\
Task description: reorganize video clips in chronological order. You need to provide 5 key events and ask to restore the timeline.
Number of questions: 1 question
Example 1: 1. The man in black wakes up 2. The man in black is attacked 3. The man in black goes out 4. The man in black is sleeping 5. The man in black meets the woman\\

Type 5: Multimodal Hint Reasoning (MHR)\\
Task description: analyze the visual and auditory hints set by the director: camera movement semantics, object position changes, sound and picture metaphors. Ask about specific camera transitions, the appearance, movement, or change of objects.
Number of questions: Create one question for each given hint shot based on the given [reasoning conclusion].
Example 1: The correct interpretation of the dancing scene between the man and woman in the video is (Answer: A. The man's own fantasy)\\

Type 6: Physical Anomaly Reasoning (PAR)\\
Task description: identify scenes in the video that do not conform to reality logic (such as magic, sci-fi elements). The first question asks about the rules of supernatural elements, and the second question asks about their implied meaning.
Number of questions: (2n questions) Create questions based on the given supernatural elements. If none are provided, skip this type.\\

Type 7: Core Theme Inference (CTI)\\
Task description: infer the core theme or deep meaning through the plot, dialogue, or symbols in the video. (The given main idea of the video is a subjective conclusion and may not be entirely correct)\\

The questions you propose need to be understood and carefully reasoned to answer. Each question needs to provide 6 options, ensuring the correct option is objectively unique (distractor options cannot express a similar meaning to the correct option, they must be clearly distinguished), and the distractor options should overlap with the correct option as much as possible (e.g., A. Because the man forgot to bring his phone B. Because the man forgot to bring his tablet C. Because the man forgot to bring his wallet). You need to provide the question type, question, answer, and explanation. Key point: The question stem cannot reveal character psychology, factual information, hint clues, and hint conclusions, viewers must find them themselves. For example: Bad question: "What does the skeleton at the end of the video imply?" - Good question: "What does the skeleton at the end of the video imply?"; Bad question: "What is the psychological state of the man when he imagines the future in the lighting change scene?" - Good question: "What is the psychological state of the man in the lighting change scene?"\\

Answer format example:\\
For each question, the 1st line: [Type]: CTI The 2nd line: [Question]: Your question; The 3rd-8th lines: A-F options; The 9th line: [Answer]: F; The 10th line: [Explanation]: Your explanation.
\end{tcolorbox}

\begin{tcolorbox}[colback=gray!5!white, colframe=gray!75!black, 
title=Model Evaluation Prompt (Thinking), boxrule=0.5mm, width=\textwidth, arc=3mm, auto outer arc]
Based on the given video, reason and answer the single-choice question. Provide your reasoning between the <think> and </think> tags, and then give your final answer between the <answer> and </answer> tags. The question is: {question}. The options are: {options}. Your answer:
\end{tcolorbox}
\begin{tcolorbox}[colback=gray!5!white, colframe=gray!75!black, 
title=Model Evaluation Prompt (Without Thinking), boxrule=0.5mm, width=\textwidth, arc=3mm, auto outer arc]
Based on the given video, answer the single-choice question. Give your final answer choise between the <answer> and </answer> tags. The question is: {question}. The options are: {options}. Your answer:
\end{tcolorbox}
\begin{tcolorbox}[colback=gray!5!white, colframe=gray!75!black, 
title=Reasoning Process Analysis Prompt (Thinking Wrong), boxrule=0.5mm, width=\textwidth, arc=3mm, auto outer arc]
You are an expert in logic. Your task is to analyze the cause of errors in a multimodal large model's responses to video reasoning tasks. The requirements are as follows:\\

I will provide you with a plot description of a short reasoning film, a question, options, the correct answer, and an explanation of the correct answer.Then I will provide you with the [thought process] of a multimodal large model and its incorrect answer. Your task is to compare the multimodal large model's thought process with the correct explanation and the film plot. Determine the most critical cause of the error from the following: \\

(1) VPE (Visual Perception Error): The model extracted incorrect visual information for analysis, leading to the wrong answer. (For example, the video shows a man robbing with a knife, but the model perceives it as a man robbing with a gun.) \\
(2) VOE (Visual Omission Error): The model failed to extract key visual information (such as key objects and events), leading to the wrong answer. \\
(3) RE (Reasoning Error): The model made an error in the reasoning process, misinterpreting the implications of the visual information and incorrectly judging the relationships between multiple pieces of visual information. \\
(4) TRAW (Think Right Answer Wrong): The model's thought process is generally consistent with the answer explanation, but it chose the wrong option when answering.\\
Your response format: Please provide the error abbreviation within <Type></Type>. Then give your brief judgment reason within <Reason></Reason>.
\end{tcolorbox}
\begin{tcolorbox}[colback=gray!5!white, colframe=gray!75!black, 
title=Reasoning Process Analysis Prompt (Thinking Right), boxrule=0.5mm, width=\textwidth, arc=3mm, auto outer arc]
You are an expert in logic. Your task is to analyze the thought process of a multimodal large model in responding to video reasoning tasks. The requirements are as follows:\\

I will provide you with a [plot description] of a short reasoning film, a question, options, the correct answer, and an explanation of the correct answer. Then I will provide you with the [model's thought process] and its correct answer. Your task is to compare the multimodal large model's thought process with the correct explanation and the film plot. Determine which type the response belongs to:\\
(1) TWAR (Think Wrong Answer Right): The model's thought process shows a significant deviation from the answer explanation, using incorrect information to arrive at the correct conclusion. \\
(2) TRAR (Think Right Answer Right): The model's thought process is generally consistent with the answer explanation (allowing for minor deviations).\\
Your response format: Please provide the response type abbreviation within <Type></Type>. Then give your brief judgment reason within <Reason></Reason>.
\end{tcolorbox}

\section{Key Statistics of Video-Holmes}
\label{app C}

\begin{wraptable}{r}{0.5\textwidth}
\vspace{-1.3em}
\caption{Key Statistics of Video-Holmes.}
\vspace{-2mm}
\label{tab: key statistics}
\small
\centering
\begin{tabular}{lc}
\toprule
\textbf{Statistic} & \textbf{Number} \\
\midrule
Total Suspense Short Films & 270 \\
\quad -  Anime & 23 (8.5\%)  \\
\quad -  Comic & 32 (11.9\%)  \\
\quad -  Detective & 31 (11.5\%)   \\
\quad -  Future & 7 (2.6\%)   \\
\quad -  Horror & 88 (32.6\%)   \\
\quad -  Social & 66 (24.4\%)   \\
\quad -  Supernatural & 36 (13.3\%)   \\
\quad -  Thriller & 23 (8.5\%)   \\
\midrule
Total Questions & 1,837  \\
\quad -  SR & 292 (15.6\%)  \\
\quad -  IMC & 276 (15.0\%)  \\
\quad -  TCI & 273 (14.9\%)   \\
\quad -  TA & 200 (10.9\%)   \\
\quad -  MHR & 332 (18.1\%)   \\
\quad -  PAR & 194 (10.4\%)   \\
\quad -  CTI & 270 (14.7\%)   \\
\midrule
Average Question Word Count  & 17.4 \\
Average Explanation Word Count & 31.3  \\
\bottomrule
\vspace{-5em}
\end{tabular}

\label{table:statistics}
\end{wraptable}
The key statistics of Video-Holmes are shown in Table~\ref{tab: key statistics}. To ensure diversity, we include nine subkeywords (Anim, Comic, Detective, Future, Horror, Social, Supernatural, Thriller) when searching for suspense short films. The distribution of the nine specifically designed tasks is relatively balanced, with a higher proportion of MHR tasks because a single video often contains more than one reasoning shot annotated by humans. PAR tasks are absent in videos without supernatural phenomena, as such questions are not applicable.

\section{Examples of Video-Holmes}
\label{app D}

\begin{figure*}[!t]
	\centering
	\includegraphics[width=\textwidth]{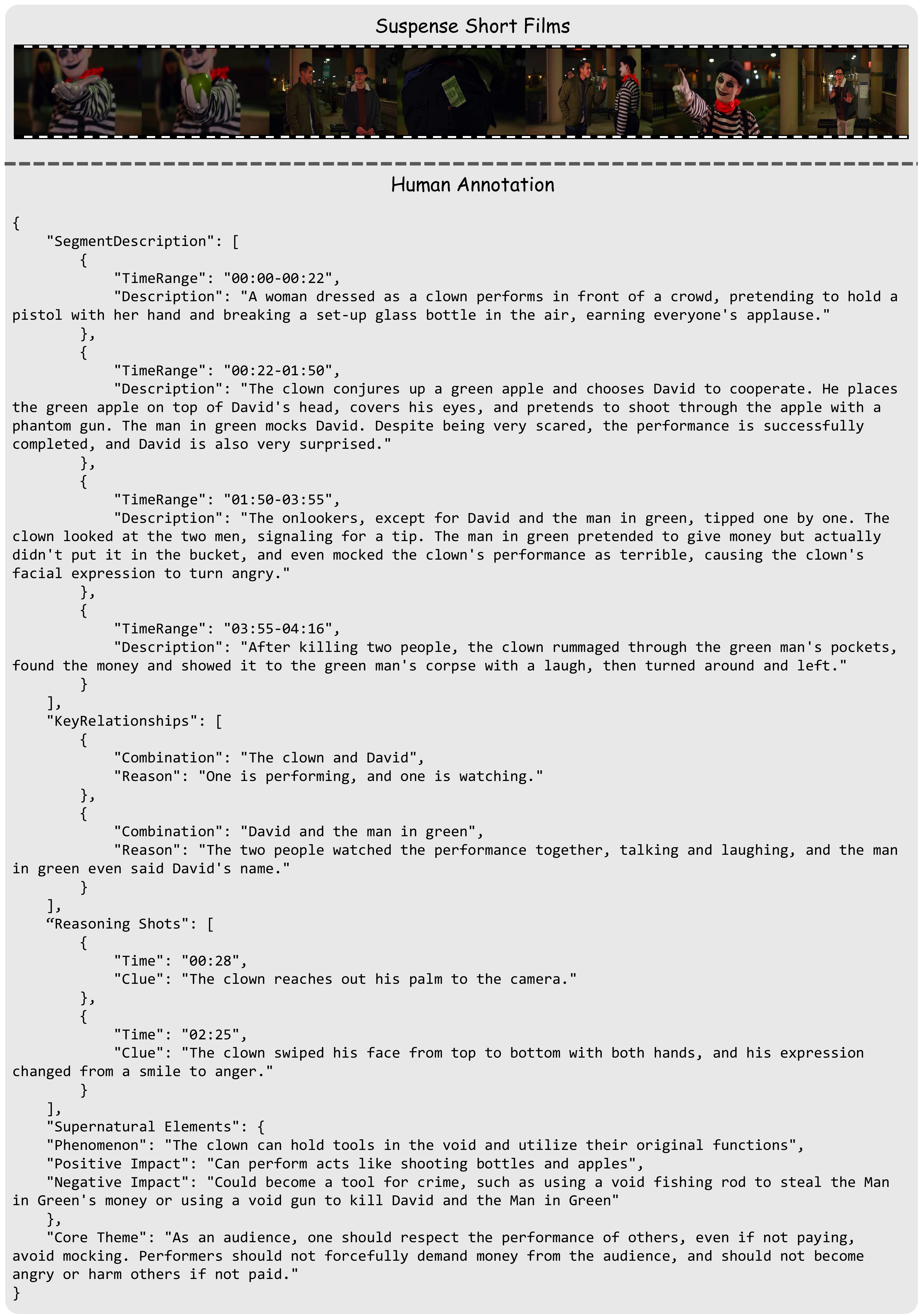}
\caption{Example of human annotation.}
\label{fig: human anno}
\end{figure*}

\begin{figure*}[!t]
	\centering
	\includegraphics[width=\textwidth]{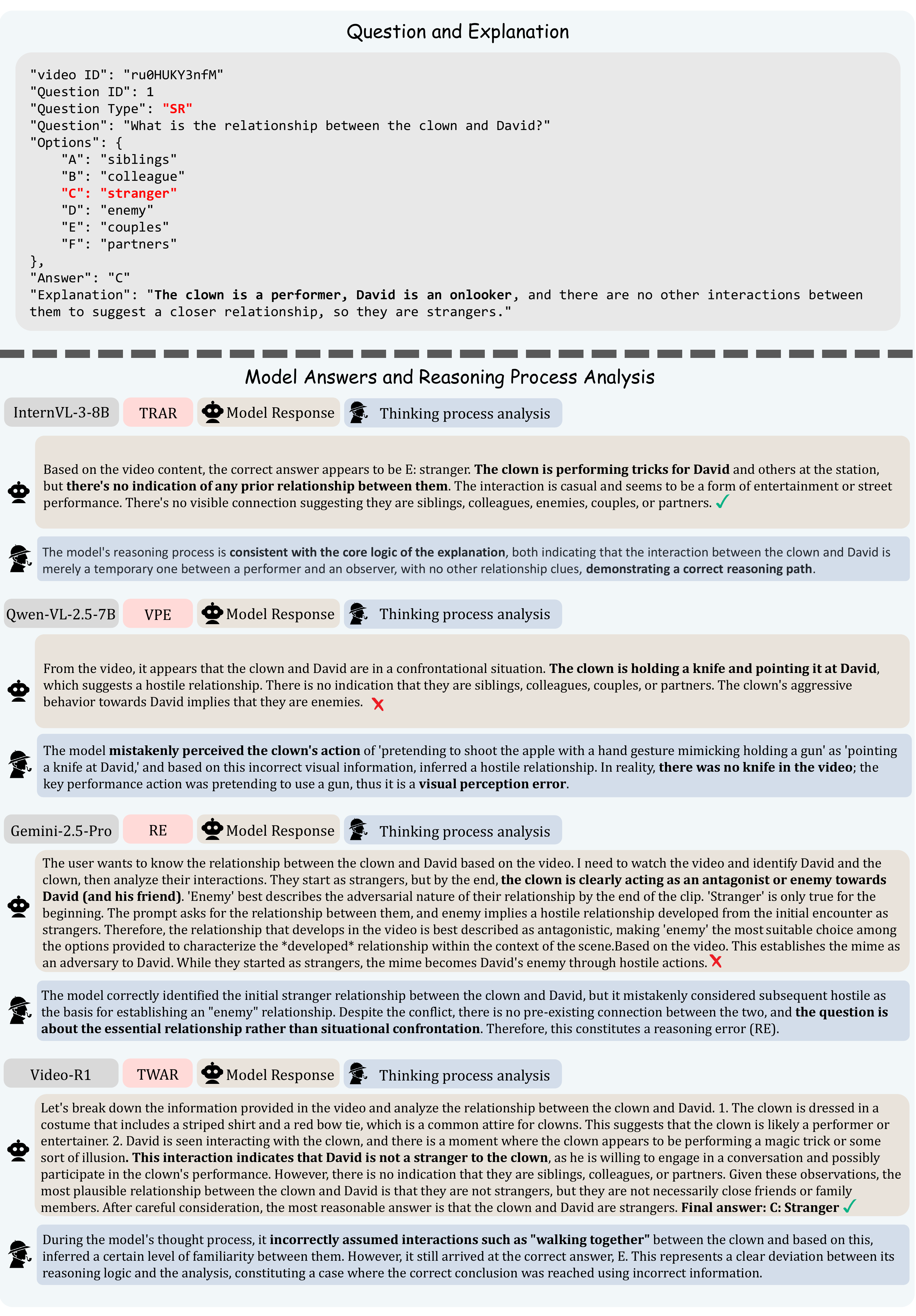}
\caption{Example of question, model answers and reasoning process analysis.}
\label{fig: q1}
\end{figure*}

\begin{figure*}[!t]
	\centering
	\includegraphics[width=\textwidth]{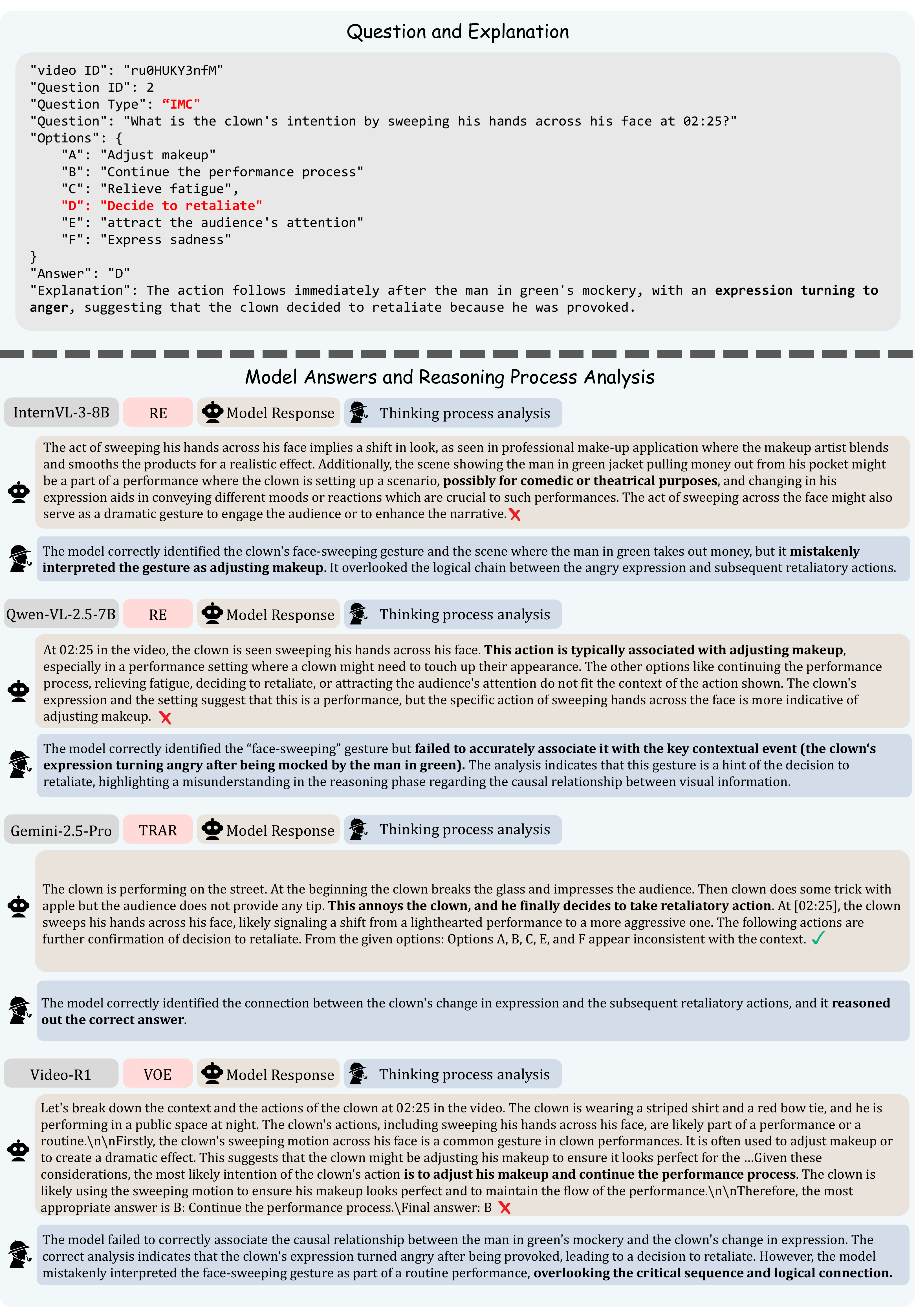}
\caption{Example of question, model answers and reasoning process analysis.}
\label{fig: q2}
\end{figure*}

\begin{figure*}[!t]
	\centering
	\includegraphics[width=\textwidth]{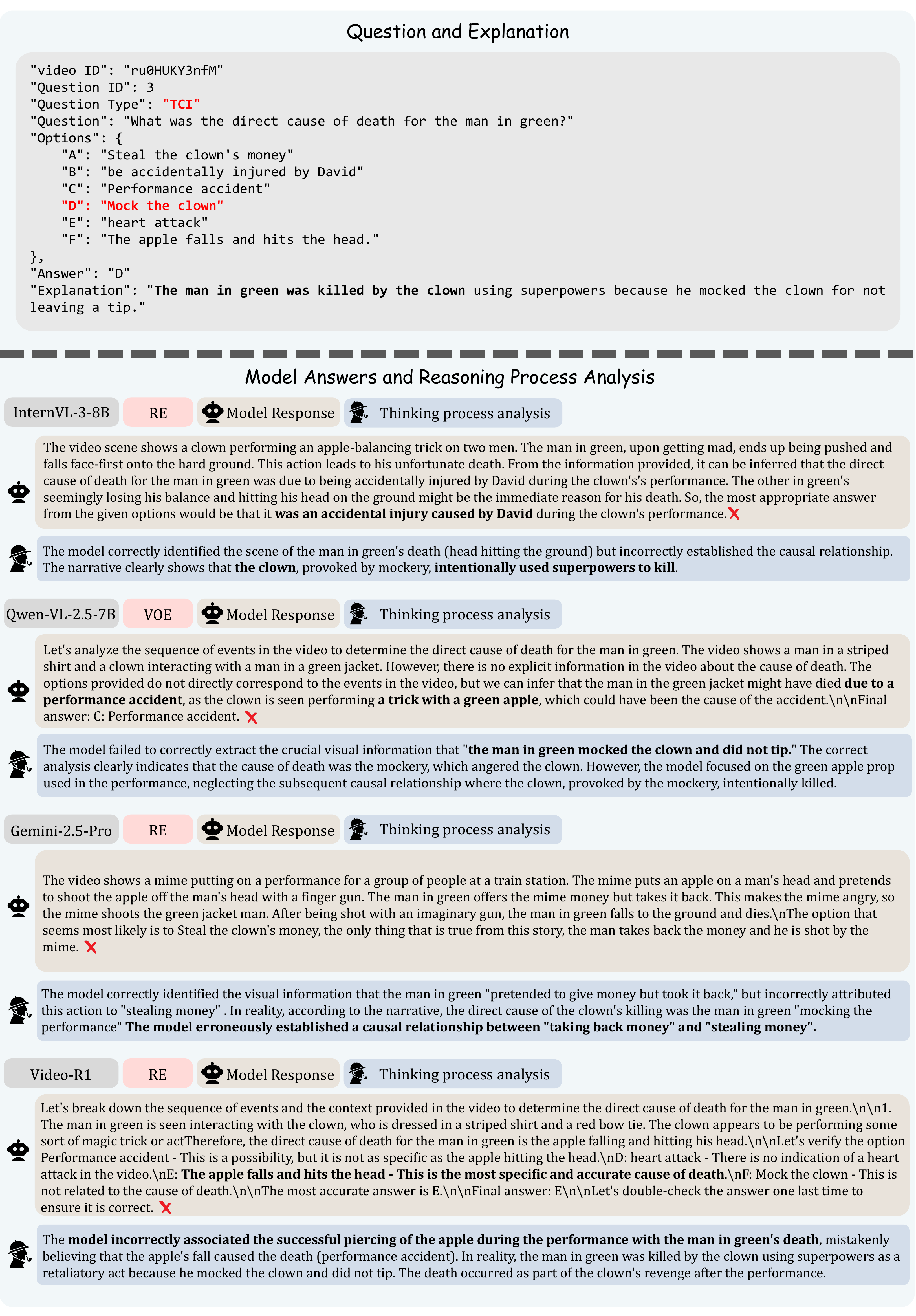}
\caption{Example of question, model answers and reasoning process analysis.}
\label{fig: q3}
\end{figure*}

\begin{figure*}[!t]
	\centering
	\includegraphics[width=\textwidth]{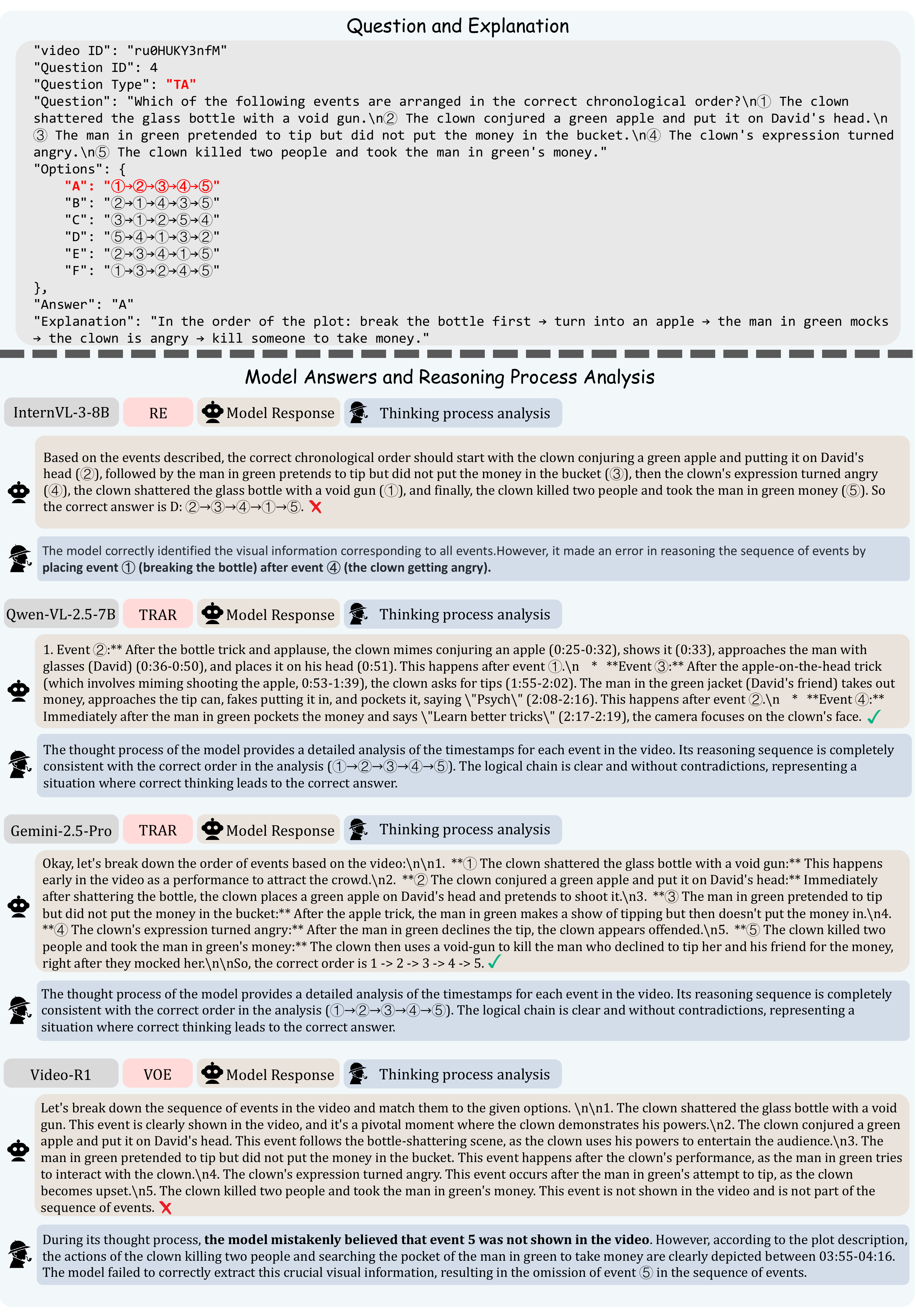}
\caption{Example of question, model answers and reasoning process analysis.}
\label{fig: q4}
\end{figure*}

\begin{figure*}[!t]
	\centering
	\includegraphics[width=\textwidth]{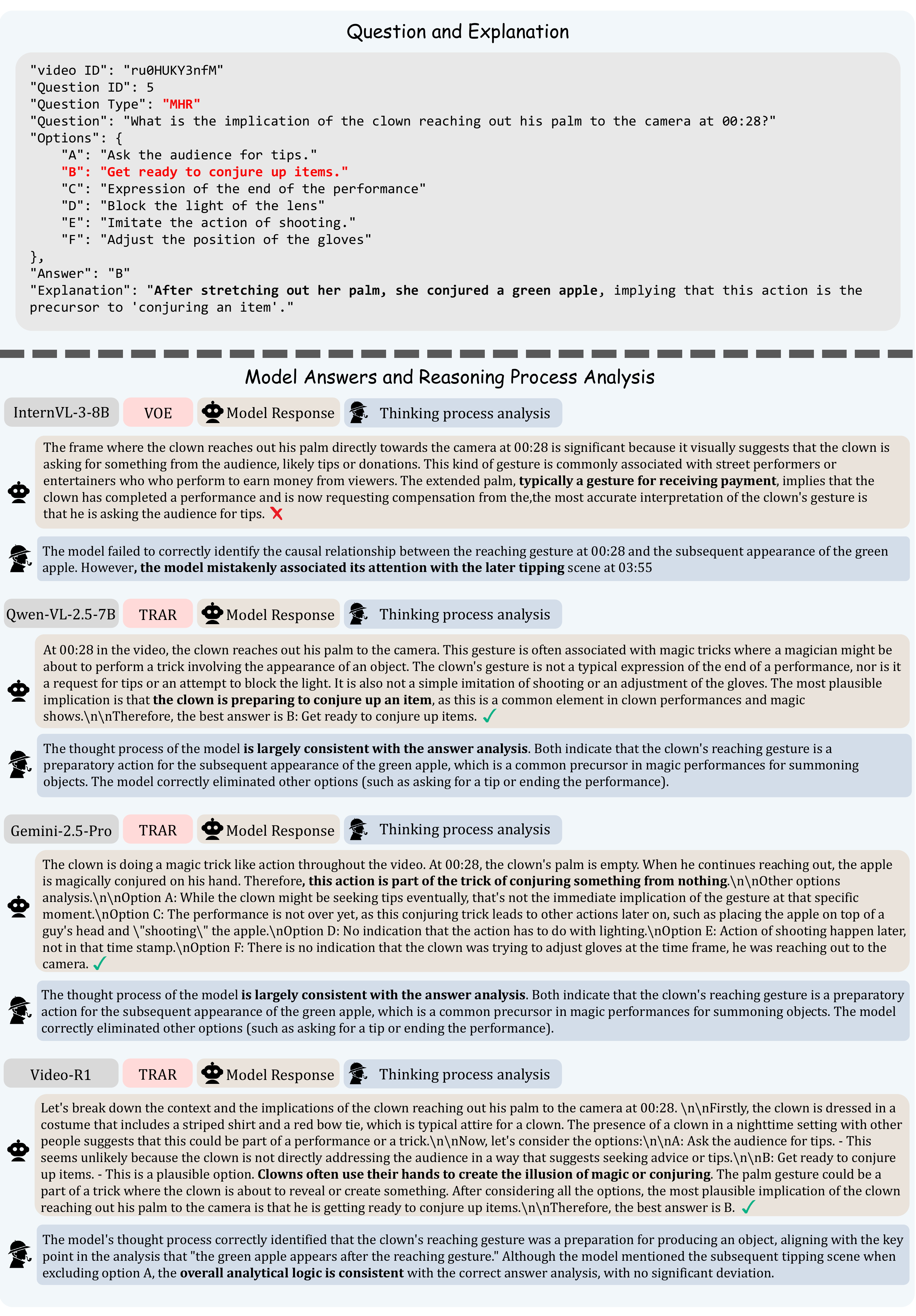}
\caption{Example of question, model answers and reasoning process analysis.}
\label{fig: q5}
\end{figure*}

\begin{figure*}[!t]
	\centering
	\includegraphics[width=\textwidth]{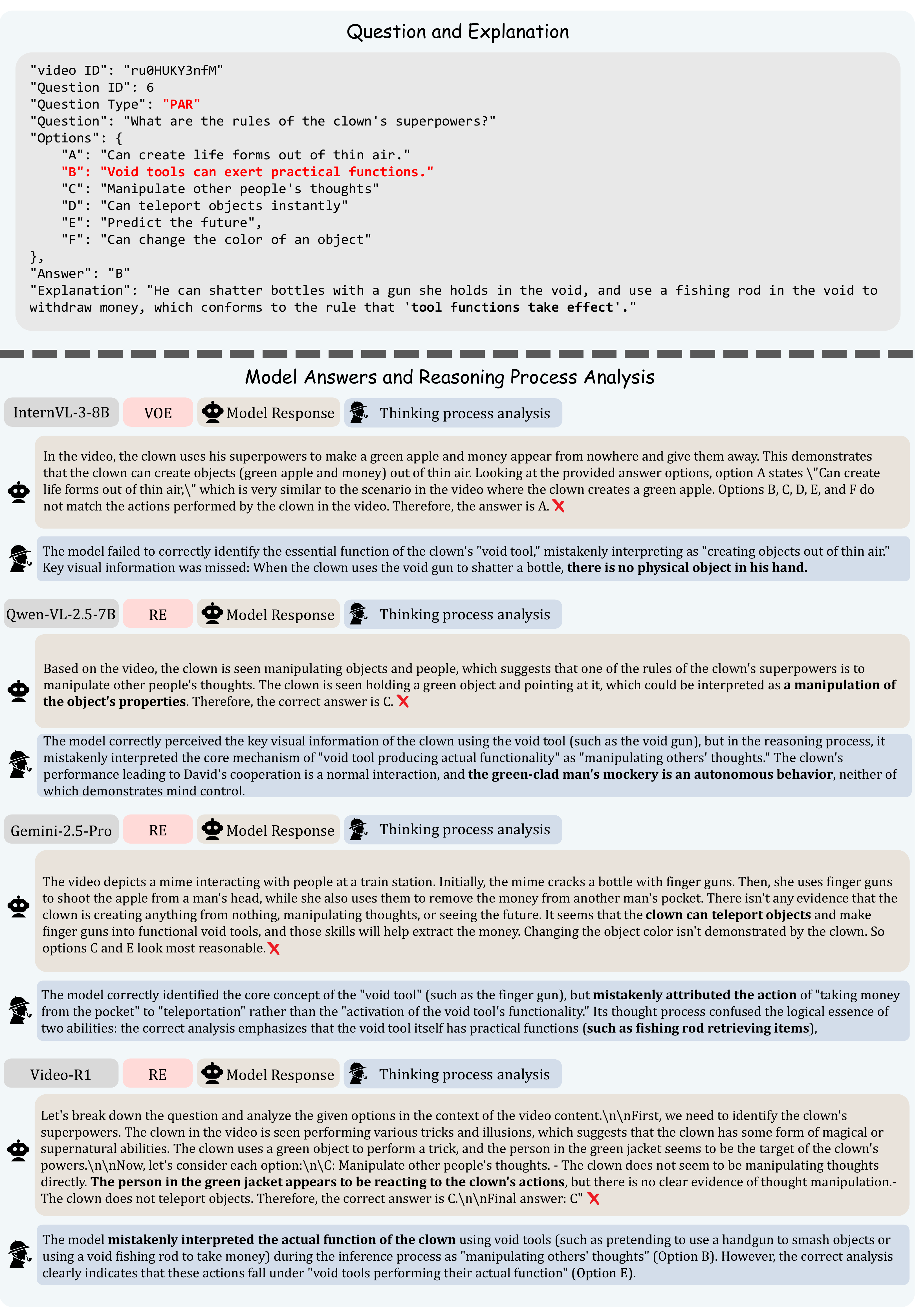}
\caption{Example of question, model answers and reasoning process analysis.}
\label{fig: q6}
\end{figure*}

\begin{figure*}[!t]
	\centering
	\includegraphics[width=\textwidth]{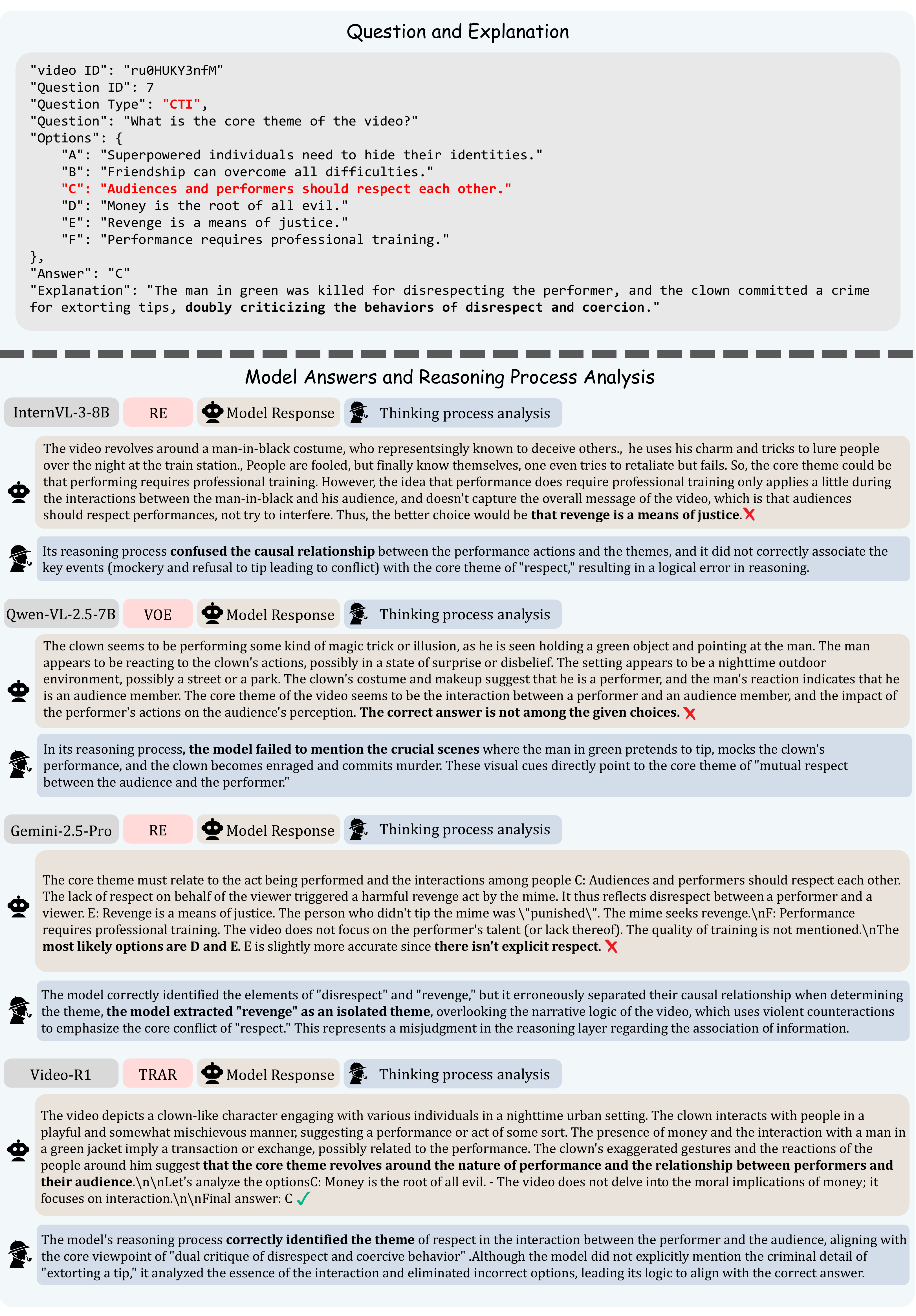}
\caption{Example of question, model answers and reasoning process analysis.}
\label{fig: q7}
\end{figure*}

Figures~\ref{fig: human anno} to \ref{fig: q7} illustrate an example of Video-Holmes. Specifically, Figure~\ref{fig: human anno} presents the human annotation results, while Figures~\ref{fig: q1} to~\ref{fig: q7} display the questions and explanations generated by DeepSeek, along with the models' answers and reasoning process analysis.

\section{Broader Impact}
\label{app E}

The development and release of the Video-Holmes benchmark have the potential to impact the field of complex video reasoning by providing a rigorous and comprehensive evaluation benchmark. However, it is important to acknowledge the potential ethical considerations associated with the use of this benchmark. The video content used in Video-Holmes, derived from suspense short films, may contain elements of horror or thriller genres, which could be distressing or inappropriate for certain audiences. Researchers and developers utilizing this benchmark should be mindful of the nature of the content and ensure that it is used responsibly, with appropriate content warnings and considerations for the intended audience.

\clearpage

\bibliographystyle{unsrt}
\bibliography{neurips_2025}

\begin{thebibliography}{10}

\bibitem{wei2022chain}
Jason Wei, Xuezhi Wang, Dale Schuurmans, Maarten Bosma, Fei Xia, Ed~Chi, Quoc~V Le, Denny Zhou, et~al.
\newblock Chain-of-thought prompting elicits reasoning in large language models.
\newblock {\em Advances in neural information processing systems}, 35:24824--24837, 2022.

\bibitem{shao2024deepseekmath}
Zhihong Shao, Peiyi Wang, Qihao Zhu, Runxin Xu, Junxiao Song, Xiao Bi, Haowei Zhang, Mingchuan Zhang, YK~Li, Y~Wu, et~al.
\newblock Deepseekmath: Pushing the limits of mathematical reasoning in open language models.
\newblock {\em arXiv preprint arXiv:2402.03300}, 2024.

\bibitem{guo2025deepseek}
Daya Guo, Dejian Yang, Haowei Zhang, Junxiao Song, Ruoyu Zhang, Runxin Xu, Qihao Zhu, Shirong Ma, Peiyi Wang, Xiao Bi, et~al.
\newblock Deepseek-r1: Incentivizing reasoning capability in llms via reinforcement learning.
\newblock {\em arXiv preprint arXiv:2501.12948}, 2025.

\bibitem{o1}
OpenAI.
\newblock Introducing openai o1.
\newblock 2024.

\bibitem{o3}
OpenAI.
\newblock Openai o3.
\newblock 2025.

\bibitem{feng2025video}
Kaituo Feng, Kaixiong Gong, Bohao Li, Zonghao Guo, Yibing Wang, Tianshuo Peng, Benyou Wang, and Xiangyu Yue.
\newblock Video-r1: Reinforcing video reasoning in mllms.
\newblock {\em arXiv preprint arXiv:2503.21776}, 2025.

\bibitem{li2025videochat}
Xinhao Li, Ziang Yan, Desen Meng, Lu~Dong, Xiangyu Zeng, Yinan He, Yali Wang, Yu~Qiao, Yi~Wang, and Limin Wang.
\newblock Videochat-r1: Enhancing spatio-temporal perception via reinforcement fine-tuning.
\newblock {\em arXiv preprint arXiv:2504.06958}, 2025.

\bibitem{chen2025exploring}
Yi~Chen, Yuying Ge, Rui Wang, Yixiao Ge, Lu~Qiu, Ying Shan, and Xihui Liu.
\newblock Exploring the effect of reinforcement learning on video understanding: Insights from seed-bench-r1.
\newblock {\em arXiv preprint arXiv:2503.24376}, 2025.

\bibitem{geminithinking}
Google.
\newblock Gemini-2.0-flash-thinking, 2024.

\bibitem{yang2024thinking}
Jihan Yang, Shusheng Yang, Anjali~W Gupta, Rilyn Han, Li~Fei-Fei, and Saining Xie.
\newblock Thinking in space: How multimodal large language models see, remember, and recall spaces.
\newblock {\em arXiv preprint arXiv:2412.14171}, 2024.

\bibitem{he2024mmworld}
Xuehai He, Weixi Feng, Kaizhi Zheng, Yujie Lu, Wanrong Zhu, Jiachen Li, Yue Fan, Jianfeng Wang, Linjie Li, Zhengyuan Yang, et~al.
\newblock Mmworld: Towards multi-discipline multi-faceted world model evaluation in videos.
\newblock {\em arXiv preprint arXiv:2406.08407}, 2024.

\bibitem{zhao2025mmvu}
Yilun Zhao, Lujing Xie, Haowei Zhang, Guo Gan, Yitao Long, Zhiyuan Hu, Tongyan Hu, Weiyuan Chen, Chuhan Li, Junyang Song, et~al.
\newblock Mmvu: Measuring expert-level multi-discipline video understanding.
\newblock {\em arXiv preprint arXiv:2501.12380}, 2025.

\bibitem{qi2025vcr}
Yukun Qi, Yiming Zhao, Yu~Zeng, Xikun Bao, Wenxuan Huang, Lin Chen, Zehui Chen, Jie Zhao, Zhongang Qi, and Feng Zhao.
\newblock Vcr-bench: A comprehensive evaluation framework for video chain-of-thought reasoning.
\newblock {\em arXiv preprint arXiv:2504.07956}, 2025.

\bibitem{hu2025video}
Kairui Hu, Penghao Wu, Fanyi Pu, Wang Xiao, Yuanhan Zhang, Xiang Yue, Bo~Li, and Ziwei Liu.
\newblock Video-mmmu: Evaluating knowledge acquisition from multi-discipline professional videos.
\newblock {\em arXiv preprint arXiv:2501.13826}, 2025.

\bibitem{li2024mvbench}
Kunchang Li, Yali Wang, Yinan He, Yizhuo Li, Yi~Wang, Yi~Liu, Zun Wang, Jilan Xu, Guo Chen, Ping Luo, et~al.
\newblock Mvbench: A comprehensive multi-modal video understanding benchmark.
\newblock In {\em Proceedings of the IEEE/CVF Conference on Computer Vision and Pattern Recognition}, pages 22195--22206, 2024.

\bibitem{liu2024tempcompass}
Yuanxin Liu, Shicheng Li, Yi~Liu, Yuxiang Wang, Shuhuai Ren, Lei Li, Sishuo Chen, Xu~Sun, and Lu~Hou.
\newblock Tempcompass: Do video llms really understand videos?
\newblock {\em arXiv preprint arXiv:2403.00476}, 2024.

\bibitem{cheng2025v}
Zixu Cheng, Jian Hu, Ziquan Liu, Chenyang Si, Wei Li, and Shaogang Gong.
\newblock V-star: Benchmarking video-llms on video spatio-temporal reasoning.
\newblock {\em arXiv preprint arXiv:2503.11495}, 2025.

\bibitem{fu2024video}
Chaoyou Fu, Yuhan Dai, Yongdong Luo, Lei Li, Shuhuai Ren, Renrui Zhang, Zihan Wang, Chenyu Zhou, Yunhang Shen, Mengdan Zhang, et~al.
\newblock Video-mme: The first-ever comprehensive evaluation benchmark of multi-modal llms in video analysis.
\newblock {\em arXiv preprint arXiv:2405.21075}, 2024.

\bibitem{zhu2023minigpt}
Deyao Zhu, Jun Chen, Xiaoqian Shen, Xiang Li, and Mohamed Elhoseiny.
\newblock Minigpt-4: Enhancing vision-language understanding with advanced large language models.
\newblock {\em arXiv preprint arXiv:2304.10592}, 2023.

\bibitem{zheng2023minigpt}
Kaizhi Zheng, Xuehai He, and Xin~Eric Wang.
\newblock Minigpt-5: Interleaved vision-and-language generation via generative vokens.
\newblock {\em arXiv preprint arXiv:2310.02239}, 2023.

\bibitem{ge2023making}
Yuying Ge, Sijie Zhao, Ziyun Zeng, Yixiao Ge, Chen Li, Xintao Wang, and Ying Shan.
\newblock Making llama see and draw with seed tokenizer.
\newblock {\em arXiv preprint arXiv:2310.01218}, 2023.

\bibitem{cao2024visdiahalbench}
Qingxing Cao, Junhao Cheng, Xiaodan Liang, and Liang Lin.
\newblock Visdiahalbench: A visual dialogue benchmark for diagnosing hallucination in large vision-language models.
\newblock In {\em Proceedings of the 62nd Annual Meeting of the Association for Computational Linguistics (Volume 1: Long Papers)}, pages 12161--12176, 2024.

\bibitem{li2023videochat}
KunChang Li, Yinan He, Yi~Wang, Yizhuo Li, Wenhai Wang, Ping Luo, Yali Wang, Limin Wang, and Yu~Qiao.
\newblock Videochat: Chat-centric video understanding.
\newblock {\em arXiv preprint arXiv:2305.06355}, 2023.

\bibitem{maaz2023video}
Muhammad Maaz, Hanoona Rasheed, Salman Khan, and Fahad~Shahbaz Khan.
\newblock Video-chatgpt: Towards detailed video understanding via large vision and language models.
\newblock {\em arXiv preprint arXiv:2306.05424}, 2023.

\bibitem{hong2024cogvlm2}
Wenyi Hong, Weihan Wang, Ming Ding, Wenmeng Yu, Qingsong Lv, Yan Wang, Yean Cheng, Shiyu Huang, Junhui Ji, Zhao Xue, et~al.
\newblock Cogvlm2: Visual language models for image and video understanding.
\newblock {\em arXiv preprint arXiv:2408.16500}, 2024.

\bibitem{chen2024internvl}
Zhe Chen, Jiannan Wu, Wenhai Wang, Weijie Su, Guo Chen, Sen Xing, Muyan Zhong, Qinglong Zhang, Xizhou Zhu, Lewei Lu, et~al.
\newblock Internvl: Scaling up vision foundation models and aligning for generic visual-linguistic tasks.
\newblock In {\em Proceedings of the IEEE/CVF conference on computer vision and pattern recognition}, pages 24185--24198, 2024.

\bibitem{zhu2025internvl3}
Jinguo Zhu, Weiyun Wang, Zhe Chen, Zhaoyang Liu, Shenglong Ye, Lixin Gu, Yuchen Duan, Hao Tian, Weijie Su, Jie Shao, et~al.
\newblock Internvl3: Exploring advanced training and test-time recipes for open-source multimodal models.
\newblock {\em arXiv preprint arXiv:2504.10479}, 2025.

\bibitem{zhang2024video}
Yuanhan Zhang, Jinming Wu, Wei Li, Bo~Li, Zejun Ma, Ziwei Liu, and Chunyuan Li.
\newblock Video instruction tuning with synthetic data.
\newblock {\em arXiv preprint arXiv:2410.02713}, 2024.

\bibitem{wang2024qwen2}
Peng Wang, Shuai Bai, Sinan Tan, Shijie Wang, Zhihao Fan, Jinze Bai, Keqin Chen, Xuejing Liu, Jialin Wang, Wenbin Ge, et~al.
\newblock Qwen2-vl: Enhancing vision-language model's perception of the world at any resolution.
\newblock {\em arXiv preprint arXiv:2409.12191}, 2024.

\bibitem{xu2025qwen2}
Jin Xu, Zhifang Guo, Jinzheng He, Hangrui Hu, Ting He, Shuai Bai, Keqin Chen, Jialin Wang, Yang Fan, Kai Dang, et~al.
\newblock Qwen2. 5-omni technical report.
\newblock {\em arXiv preprint arXiv:2503.20215}, 2025.

\bibitem{team2025kimi}
Kimi Team, Angang Du, Bohong Yin, Bowei Xing, Bowen Qu, Bowen Wang, Cheng Chen, Chenlin Zhang, Chenzhuang Du, Chu Wei, et~al.
\newblock Kimi-vl technical report.
\newblock {\em arXiv preprint arXiv:2504.07491}, 2025.

\bibitem{liao2025improved}
Zhenyi Liao, Qingsong Xie, Yanhao Zhang, Zijian Kong, Haonan Lu, Zhenyu Yang, and Zhijie Deng.
\newblock Improved visual-spatial reasoning via r1-zero-like training.
\newblock {\em arXiv preprint arXiv:2504.00883}, 2025.

\bibitem{deng2025openvlthinker}
Yihe Deng, Hritik Bansal, Fan Yin, Nanyun Peng, Wei Wang, and Kai-Wei Chang.
\newblock Openvlthinker: An early exploration to complex vision-language reasoning via iterative self-improvement.
\newblock {\em arXiv preprint arXiv:2503.17352}, 2025.

\bibitem{shen2025vlm}
Haozhan Shen, Peng Liu, Jingcheng Li, Chunxin Fang, Yibo Ma, Jiajia Liao, Qiaoli Shen, Zilun Zhang, Kangjia Zhao, Qianqian Zhang, et~al.
\newblock Vlm-r1: A stable and generalizable r1-style large vision-language model.
\newblock {\em arXiv preprint arXiv:2504.07615}, 2025.

\bibitem{xu2024llava}
Guowei Xu, Peng Jin, Li~Hao, Yibing Song, Lichao Sun, and Li~Yuan.
\newblock Llava-o1: Let vision language models reason step-by-step.
\newblock {\em arXiv preprint arXiv:2411.10440}, 2024.

\bibitem{thawakar2025llamav}
Omkar Thawakar, Dinura Dissanayake, Ketan More, Ritesh Thawkar, Ahmed Heakl, Noor Ahsan, Yuhao Li, Mohammed Zumri, Jean Lahoud, Rao~Muhammad Anwer, et~al.
\newblock Llamav-o1: Rethinking step-by-step visual reasoning in llms.
\newblock {\em arXiv preprint arXiv:2501.06186}, 2025.

\bibitem{zhang2025tinyllava}
Xingjian Zhang, Siwei Wen, Wenjun Wu, and Lei Huang.
\newblock Tinyllava-video-r1: Towards smaller lmms for video reasoning.
\newblock {\em arXiv preprint arXiv:2504.09641}, 2025.

\bibitem{wang2025timezero}
Ye~Wang, Boshen Xu, Zihao Yue, Zihan Xiao, Ziheng Wang, Liang Zhang, Dingyi Yang, Wenxuan Wang, and Qin Jin.
\newblock Timezero: Temporal video grounding with reasoning-guided lvlm.
\newblock {\em arXiv preprint arXiv:2503.13377}, 2025.

\bibitem{bai2025qwen2}
Shuai Bai, Keqin Chen, Xuejing Liu, Jialin Wang, Wenbin Ge, Sibo Song, Kai Dang, Peng Wang, Shijie Wang, Jun Tang, et~al.
\newblock Qwen2. 5-vl technical report.
\newblock {\em arXiv preprint arXiv:2502.13923}, 2025.

\bibitem{xu2017video}
Dejing Xu, Zhou Zhao, Jun Xiao, Fei Wu, Hanwang Zhang, Xiangnan He, and Yueting Zhuang.
\newblock Video question answering via gradually refined attention over appearance and motion.
\newblock In {\em ACM Multimedia}, 2017.

\bibitem{yu2019activitynet}
Zhou Yu, Dejing Xu, Jun Yu, Ting Yu, Zhou Zhao, Yueting Zhuang, and Dacheng Tao.
\newblock Activitynet-qa: A dataset for understanding complex web videos via question answering.
\newblock In {\em Proceedings of the AAAI Conference on Artificial Intelligence}, volume~33, pages 9127--9134, 2019.

\bibitem{xiao2021next}
Junbin Xiao, Xindi Shang, Angela Yao, and Tat-Seng Chua.
\newblock Next-qa: Next phase of question-answering to explaining temporal actions.
\newblock In {\em Proceedings of the IEEE/CVF conference on computer vision and pattern recognition}, pages 9777--9786, 2021.

\bibitem{xu2023mmbench}
Cheng Xu, Xiaofeng Hou, Jiacheng Liu, Chao Li, Tianhao Huang, Xiaozhi Zhu, Mo~Niu, Lingyu Sun, Peng Tang, Tongqiao Xu, et~al.
\newblock Mmbench: Benchmarking end-to-end multi-modal dnns and understanding their hardware-software implications.
\newblock In {\em 2023 IEEE International Symposium on Workload Characterization (IISWC)}, pages 154--166. IEEE, 2023.

\bibitem{wu2024longvideobench}
Haoning Wu, Dongxu Li, Bei Chen, and Junnan Li.
\newblock Longvideobench: A benchmark for long-context interleaved video-language understanding.
\newblock {\em Advances in Neural Information Processing Systems}, 37:28828--28857, 2024.

\bibitem{chen2024expanding}
Zhe Chen, Weiyun Wang, Yue Cao, Yangzhou Liu, Zhangwei Gao, Erfei Cui, Jinguo Zhu, Shenglong Ye, Hao Tian, Zhaoyang Liu, et~al.
\newblock Expanding performance boundaries of open-source multimodal models with model, data, and test-time scaling.
\newblock {\em arXiv preprint arXiv:2412.05271}, 2024.

\bibitem{pichai2024introducing}
Sundar Pichai, D~Hassabis, and K~Kavukcuoglu.
\newblock Introducing gemini 2.0: our new ai model for the agentic era, 2024.

\bibitem{gemini2}
Google.
\newblock Gemini-2.0-pro, 2025.

\bibitem{gemini25pro}
Google.
\newblock Gemini-2.5-pro, 2025.

\bibitem{4o}
OpenAI.
\newblock Hello gpt-4o, 2024.

\bibitem{o4mini}
OpenAI.
\newblock o4-mini, 2025.

\bibitem{claud}
Anthropic.
\newblock The claude 3 model family: Opus, sonnet, haiku, 2024.

\end{thebibliography}

\end{document}